\documentclass[runningheads]{llncs}

\usepackage{eccv}

\usepackage{eccvabbrv}

\usepackage{hyperref}       % hyperlinks
\usepackage{url}            % simple URL typesetting
\usepackage{booktabs}       % professional-quality tables
\usepackage{amsfonts}       % blackboard math symbols
\usepackage{nicefrac}       % compact symbols for 1/2, etc.
\usepackage{microtype}      % microtypography
\usepackage{xcolor}         % colors
\usepackage{wrapfig}
\usepackage{comment}
\usepackage{amsmath,amssymb}
\usepackage{arydshln}
\usepackage{bbm}
\usepackage{graphicx}
 \usepackage{multirow}
\usepackage{graphicx}
\usepackage{tabularx}
\usepackage{enumitem} 
\usepackage{caption}
\usepackage{comment}
\usepackage{thm-restate}
\usepackage{ragged2e}
\usepackage{threeparttable} 
\usepackage{pifont}     
\usepackage{placeins}
\usepackage{makecell}
\usepackage{capt-of}

\usepackage[accsupp]{axessibility}  % Improves PDF readability for those with disabilities.
\newcommand{\cmark}{\ding{51}}%
\newcommand{\xmark}{\ding{55}}%

\usepackage{hyperref}

\usepackage{orcidlink}
\newcommand{\stdv}[1]{\textcolor{gray}{\scriptsize\,$\pm$#1}}

\begin{document}

% ---------------------------------------------------------------
% TODO REVIEW: Replace with your title
\title{Improving Knowledge Distillation Under Unknown Covariate Shift Through Confidence-Guided Data Augmentation} 

% TODO REVIEW: If the paper title is too long for the running head, you can set
% an abbreviated paper title here. If not, comment out.
\titlerunning{ConfiG}

% TODO FINAL: Replace with your author list. 
% Include the authors' OCRID for the camera-ready version, if at all possible.
\author{Niclas Popp \inst{1,2} \and
Kevin Alexander Laube \inst{1} \and
Matthias Hein \thanks{Joint senior author.} \inst{2} \and
Lukas Schott\thanks{Joint senior author, work done while at Bosch Center for Artificial Intelligence.} \inst{3}}

% TODO FINAL: Replace with an abbreviated list of authors.
\authorrunning{N.~Popp et al.}
% First names are abbreviated in the running head.
% If there are more than two authors, 'et al.' is used.

% TODO FINAL: Replace with your institution list.
\institute{ Bosch Center for Artificial Intelligence \\ \texttt{\{niclas.popp, kevinalexander.laube\}@de.bosch.com} \and T{\"u}bingen AI Center, University of Tübingen \\ \texttt{matthias.hein@uni-tuebingen.de}
 \and
Aleph Alpha \\ \texttt{lukas.schott@aleph-alpha.com}}

\maketitle

\begin{abstract}
    Large foundation models trained on extensive datasets demonstrate strong zero-shot capabilities in various domains.
    Knowledge distillation has become an established tool for transferring knowledge from foundation models to small student networks when data and model size are constrained.
    However, the efficacy of distillation is often hampered by limited training data coverage. This can result in a covariate shift between training and test data which in turn can lead the student to exploit spurious features or even shortcut learning.
    We address this problem by introducing a novel diffusion-based data augmentation strategy that generates images by maximizing the disagreement between the teacher and the student, effectively creating challenging samples that the student struggles with, thus mitigating the problem of covariate shift.
    Experiments demonstrate that, compared to state-of-the-art diffusion-based data augmentation baselines, our approach is best or second-best in sample mean accuracy and improves the worst group and mean group accuracy on CelebA-HQ, SpuCo Birds and BAR as well as the spurious score on Spurious ImageNet under covariate shift.
\end{abstract}

\section{Introduction}

\begin{figure}[t]
    \centering
    \includegraphics[width=1.0\columnwidth]{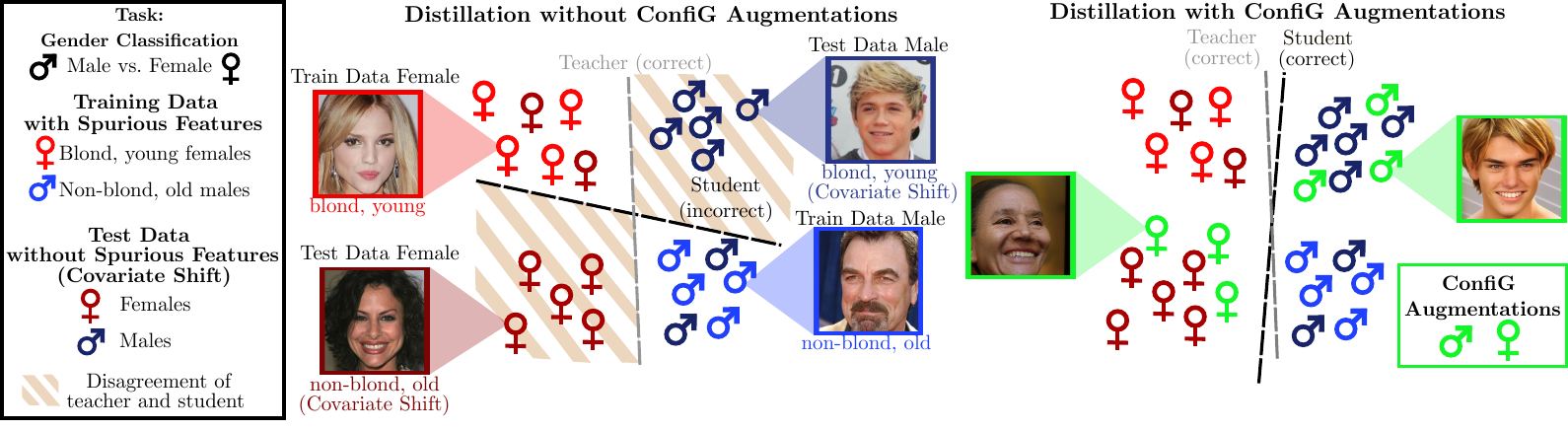}
    \caption{\textbf{Knowledge Distillation under Covariate Shift with ConfiG Data Augmentations.}
    Our goal is to maximize student performance even on groups that are fully absent from the training data.
    \textbf{Left:} Without having seen similar samples, the student is incapable of correctly classifying the unseen groups in the test set.
    \textbf{Right:} Leveraging the discrepancy between a robust teacher and the biased student, ConfiG generates and adds challenging samples to the training set.
    Further training improves the student's performance on previously missing groups by removing shortcuts.}
    \label{fig:teaser}
    \vspace{-.5cm}
\end{figure}

Foundation models have demonstrated strong zero-shot capabilities and distributional robustness across a wide range of data domains~\cite{clip,open_clip}. However, training these models demands massive datasets and computational resources, and their deployment in resource-constrained environments is often infeasible due to their scale. Achieving similar generalization and robustness with substantially smaller models and less training data remains an open research area.
Knowledge distillation (KD)~\cite{KD} offers a way to transfer knowledge from a pre-trained teacher model to a smaller student model.
Compared to supervised training with only the hard ground-truth labels, distillation has been shown to improve both performance and generalization of the student model~\cite{VanillaKD,statistical,pmlr-v97-phuong19a}.
The reason why knowledge distillation is believed to improve generalization is that the output of the teacher contains more information than hard labels (referred to as ``dark knowledge'').
However, the availability of task-specific training data is often a crucial limitation on the knowledge that can be transferred to the student.
Especially in the limited-data regime, the training data might not sufficiently cover the data distribution that the student will face during deployment.
\\
In this work, we investigate the challenge of covariate shift between the training and test data distributions in knowledge distillation.
Covariate shift refers to a situation where the distribution of the input features (covariates) changes between the training and test datasets while the conditional distribution of the output given the input remains the same.
In particular, covariate shift between training and test data can lead to ``spurious features'' or even severe ``shortcut learning''~\cite{shortcut} which enables correct decision making on the training but not on the test data.
An example for this setting is illustrated in \cref{fig:teaser}.
The task is gender classification on a training data subset of CelebA-HQ, which only contains young and blond females as well as non-blond, old males.
At test time, however, the student also encounters non-blond females or blond males.
Without training data from these groups, the student fails to learn the correct decision boundary but instead relies on spurious features such as hair color or age.
Assuming the pre-trained teacher disentangles spuriously correlated covariates and features that are predictive for the target classes, our goal is to distill a student that is similarly capable.

\noindent Several recent works improve model performance in the limited-data regime through diffusion-based data augmentations~\cite{real_guidance,gid,distdiff,dafusion,Diff-Mix,diff-ii,actgen}.
These approaches can be interpreted as a form of implicit distillation, where a generative model is used to provide additional knowledge in the form of augmented training samples.
However, by only looking at existing training samples or the student model, the augmented samples cannot effectively target the student's reliance on spurious features.
To this end, we introduce ConfiG, a \underline{Confi}dence-\underline{G}uided diffusion-based data augmentation method that leverages the disagreement between teacher and student to identify and mitigate student biases without requiring explicit knowledge of what these biases are.
As illustrated in \cref{fig:teaser}, ConfiG generates synthetic data augmentations in the regions of disagreement (dashed), such as women with brown hair or blond, young men. 
Further training on the real training images together with the augmented samples then leads the student to better approximate the decision function of the teacher, reducing the reliance on spurious features.
Our contributions are summarized as follows:

\setlist{nosep}
\begin{enumerate}[leftmargin=*,noitemsep]
    \item We demonstrate the problem that, while knowledge distillation from a robust teacher improves overall student performance, under unknown covariate shift the student still performs poorly on test samples from unseen groups.
    \item We introduce ConfiG, a data augmentation framework for knowledge distillation that guides a diffusion model to generate samples that target covariate shift through maximizing teacher and student disagreement.
    \item We provide empirical and theoretical motivation for ConfiG and demonstrate its superior performance compared to prior diffusion-based augmentations under unknown covariate shift in knowledge distillation on CelebA-HQ~\cite{liu2015faceattributes,celebahq}, SpuCo Birds~\cite{joshi2025challengesopportunitiesimprovingworstgroup}, BAR~\cite{lff}, and a subset of ImageNet~\cite{imagenet,spurious_imagenet}.
\end{enumerate}

\vspace{-.1cm}
\section{Related Work}
\vspace{-.1cm}
\textbf{Knowledge distillation} (KD) is a widely used training framework ~\cite{survery2021,survery2025} to transfer knowledge from one model (teacher) to another (student).
In contrast to empirical risk minimization, in pure response-based KD the student is trained to match the predictive distribution of the teacher~\cite{survery2025}.
If the teacher is sufficiently accurate, doing so provably improves the generalization of the student over training with hard labels~\cite{statistical,does_kd_work,feature_dist,cutmixpick}.
While there exist a lot of KD variants, simply matching the responses of student and teacher~\cite{KD} has been shown to be highly effective for image classification~\cite{patientandconsistent,VanillaKD}.
We adopt response-based knowledge distillation purely based on the soft labels of the teacher and investigate the influence of the training data.\\
\textbf{Synthetic data augmentation}
is a rising trend to mitigate data scarcity and bias.
Unlike standard augmentations like CutMix~\cite{cutmix} or MixUp~\cite{mixup}, generative models like GANs~\cite{DatasetGAN,BigDatasetGAN,awada,dad} or diffusion models can perform semantic augmentation, e.g. changing the background or attributes like hair color~\cite{freemask,syntheticdatafromdiff,lumen,aldm,clever_hans,aspire,SaSPA,datadream,bob,sgd_mix,HiGFA,salient}.
This can be viewed as indirect knowledge distillation, where the generated data represents the implicit knowledge of the generative teacher.
In our work, we combine this implicit knowledge distillation together with explicit knowledge distillation by guiding the augmentation process based on the confidences of teacher and student.
The assumption of also having an explicit teacher for the classification task is usually not stronger in practice, as models like CLIP~\cite{clip,open_clip} are trained on similar datasets as diffusion models.\\
\textbf{Group Robustness without Annotations} Common methods for achieving group robustness rely on labels for majority and minority groups in either the training~\cite{distr_robust,llrt,fairness_survey,fairness_survey_2} or validation data~\cite{jtt,llrtfewer}.
Recent approaches improve group robustness without requiring explicit group labels~\cite{effbias,ula,afr,lff,ensemble_dist}, but require all relevant groups to be represented in the dataset.
This assumption is still suboptimal in practice as relevant groups may be \textit{unknowingly absent}.
Our proposed method is designed to improve robustness even for groups that are not represented at all.

\section{Methods}
We introduce ConfiG, a \underline{Confi}dence-\underline{G}uided data augmentation method for knowledge distillation.
We first review the concept of empirical distilled risk minimization for knowledge distillation and subsequently describe the methodology of ConfiG based on guided diffusion.

\subsection{Background: Empirical Risk and Empirical Distilled Risk}\label{sec:background}
We consider classification with training data $\mathcal{D}_{\text{train}} = \{(\mathbf{x}_i, y_i)\}_{i=1}^N$ sampled from a distribution $\mathcal{P}_{\text{train}}$ and test data $\mathcal{D}_{\text{test}} \sim \mathcal{P}_{\text{test}}$.
Both share a common discrete label space $[L]$. A student model $\mathbf{f}$ and a teacher model $\mathbf{t}$ produce predictions in form of probability distributions over the label space. For a given input $\mathbf{x}$ with ground-truth label $y$, we denote the predicted probability assigned to the correct class $y$ by the student and teacher as $\mathbf{f}(\mathbf{x})_y$ and $\mathbf{t}(\mathbf{x})_y$ respectively. We refer to these values as the \emph{confidences}. 
The true test risk is defined as 
\begin{equation}
\begin{split}
    \mathcal{R}_{\mathcal{P}_\text{test}} (\textbf{f})
    & = \mathbb{E}_{(\textbf{x},y)\sim\mathcal{P}_{\text{test}}}[\ell(\mathbf{f}(\mathbf{x}),y)] 
     = \mathbb{E}_{x\sim\mathcal{P}_{\text{test},\mathbf{x}}}[\mathbb{E}_{y|\textbf{x}}[\ell(\mathbf{f}(\mathbf{x}),y)]] \\
    & = \mathbb{E}_{\mathcal{P}_{\text{test},\mathbf{x}}}[\mathbf{p}^*(\mathbf{x})^\top \mathbf{l}(\mathbf{f}(\mathbf{x}))]
\end{split}
\end{equation}
where $\mathbf{p}^*(\mathbf{x}) = [\mathbb{P}(y|\mathbf{x})]_{y\in[L]}$ is the data-generating conditional distribution over the labels, and $\mathbf{l}(\mathbf{f}(\mathbf{x}))=[\ell(\mathbf{f}(\mathbf{x}),y)]_{y\in[L]}$ denotes the vector of losses for each possible label. Similarly, the true train risk is 
\begin{equation}
    \mathcal{R}_{\text{train}}(\textbf{f}) = \mathbb{E}_{(x,y)\sim\mathcal{P}_{\text{train}}}[\ell(\mathbf{f}(\mathbf{x}),y)] = \mathbb{E}_{\mathcal{P}_{\text{train,x}}}[\mathbf{p}^*(\mathbf{x})^\top \mathbf{l}(\mathbf{f}(\mathbf{x}))].
\end{equation}
In practice, the true data distribution is unknown, and models are instead trained by minimizing the empirical risk on the training data. In standard empirical risk minimization (\textbf{ERM}) the student model $\mathbf{f}$ is trained to minimize the empirical approximation of $\mathcal{R}_{\text{train}}$ using the labels of the train samples. The empirical risk on $\mathcal{D}_{\text{train}}$ is defined as  
\begin{equation}
R_{\text{train}}(\textbf{f}) = \frac{1}{N} \sum_{i=1}^N \ell(\mathbf{f}(\mathbf{x}_i), y_i) = -\frac{1}{N} \sum_{i=1}^N \textbf{e}_{y_i}^T  \log(\mathbf{f}(\mathbf{x}_i)),
\end{equation}
where we choose $\ell(\cdot, \cdot)$ to be  the cross-entropy loss.
In knowledge distillation, the student is trained by empirical distilled risk minimization (\textbf{EDRM}) which optimizes the cross-entropy loss between the teacher and student predictive probabilities:
\begin{equation}
R_{\text{train}}^D(\textbf{f}) = -\frac{1}{N} \sum_{i=1}^N \mathbf{t}(\mathbf{x}_i)^\top \log(\mathbf{f}(\mathbf{x}_i)).
\end{equation}
Here, $\mathbf{t}(\mathbf{x}_i)$ refers to the vector with the predicted probability for each class by the teacher at $\textbf{x}_i$. The student is trained to match these soft targets provided by the teacher instead of the hard labels. The generalization error of the student under this training scheme is defined as 
\begin{equation}
    \Delta = \mathcal{R}_{\text{test}}(\textbf{f}) - R_{\text{train}}^D(\textbf{f}).
\end{equation}
Using a teacher which approximates $\mathbf{p}^*$ closely has been shown both theoretically and empirically to improve the generalization error if $\mathcal{P}_{\text{train}} = \mathcal{P}_{\text{test}}$~\cite{statistical,cutmixpick,efficient_kd,pmlr-v97-phuong19a}. This corresponds to the setting \textit{without} covariate shift where the input distribution for training and test data is the same. In the case of covariate shift, the input marginal distributions $\mathcal{P}_{\text{train},\mathbf{x}}$ and $\mathcal{P}_{\text{test},\mathbf{x}}$ are different and only the data-generating conditional distribution $\textbf{p}^*$ is the same at train and test time.

\subsection{Confidence-Guided Data Augmentations using Diffusion Models}\label{sec:config}
\begin{wrapfigure}{r}{0.5\textwidth}
    \centering
    \vskip-0.9cm
    \includegraphics[width=.5\columnwidth]{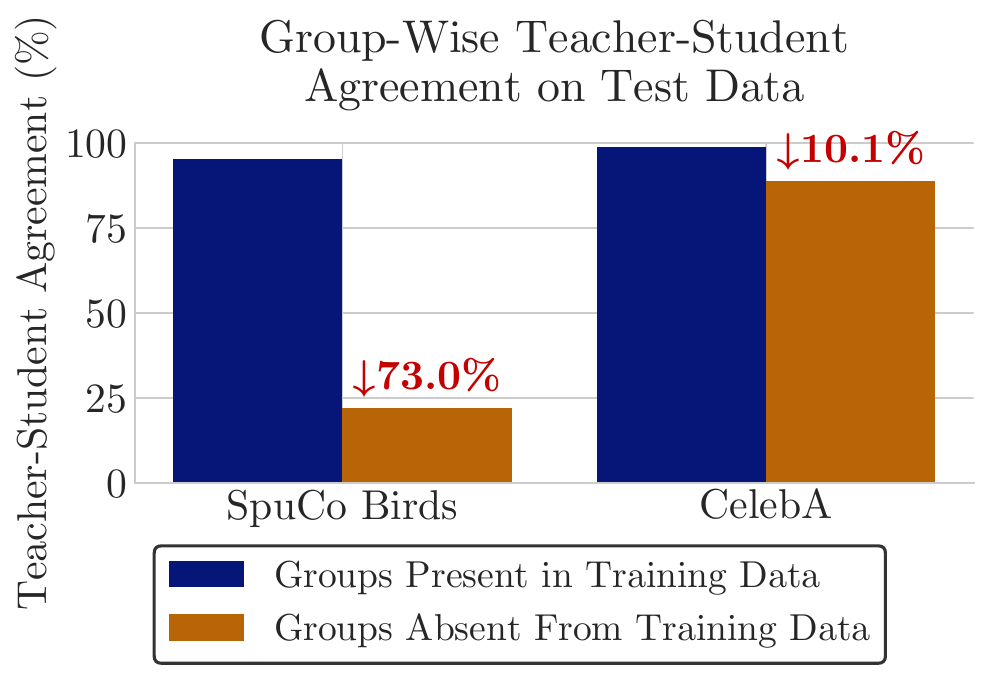}
    \vskip-0.1cm
    \caption{For a student distilled only on the biased real training data (see \cref{sec:experiments}), the agreement between teacher and student is lower on groups which are not present in the training data.
    }
    \label{fig:fidelity}

    \vskip-0.42cm
    \begin{threeparttable}
    \scriptsize 
    \renewcommand{\theadfont}{\scriptsize}
    \captionsetup{type=table}
    \caption{The vast majority of test images where student and teacher disagree belong to groups not encountered in training. The students are the same as in \cref{fig:fidelity}}
    \label{tab:disagreement}
    \begin{tabularx}{\linewidth}{@{}l|c|c|c@{}}
    \toprule
    \multicolumn{4}{c}{\footnotesize   Group-Wise Disagreements on Test Data} \\
    \midrule
    \thead{Data- \\ set} & \thead{Total Dis-\\agreements} & \thead{On Groups \\in Trainset} & \thead{On Groups Not\\in Trainset} \\
    \midrule
    CelebA & 2351 & 95 (\textbf{4\%})   & 2256 (\textbf{96\%}) \\
    SpuCo & 828  & 49 (\textbf{6\%})   & 779 (\textbf{94\%})  \\
    \bottomrule
    \end{tabularx}
    \end{threeparttable}
    \vskip-0.2cm
\end{wrapfigure}
For knowledge distillation under covariate shift, we consider the case where $\mathcal{P}_{\text{train}} \neq \mathcal{P}_{\text{test}}$.
While the error between the distilled risk and the true train risk $\mathcal{R}_{\text{train}} - R_{\text{train}}^D$ has been studied in prior works~\cite{statistical,cutmixpick}, in the case of covariate shift this does not correspond to the true  generalization error.
Therefore, we first decompose $\Delta$ into two terms
\begin{equation}
    \Delta = \Omega + \Psi
\end{equation}
where $\Psi=\mathcal{R}_{\text{train}} (\textbf{f}) - R_{\text{train}}^D (\textbf{f})$ and  $\Omega = \mathcal{R}_{\text{test}}(\textbf{f}) - \mathcal{R}_{\text{train}}(\textbf{f})$. While $\Psi$ depends on how well the teacher approximates $\mathbf{p}^*$ on $\mathcal{P}_{\text{train}}$~\cite{statistical,mse_dist}, $\Omega$ depends on the distribution of the training and test data. The goal of our data augmentation scheme is to reduce $|\Omega|$. \newline
\vspace{-.4cm}

\noindent\textbf{Empirical Motivation.} Teacher-Student Disagreement under Covariate Shift We assume access to a teacher that approximates $\mathbf{p}^*(\mathbf{x})$ well on both $\mathcal{P}_{\text{train}}$ and $\mathcal{P}_{\text{test}}$.
In \cref{fig:fidelity}, we distill a student on training data where certain groups from the test data are fully absent and analyze the agreement of teacher and student on test data. 
\noindent For instance, in CelebA-HQ, where the task is gender classification (female/male), our training data only includes female samples that are young and blond.
The test data, however, also includes females who are older or non-blond.
We observe that the agreement between teacher and student is high for groups that are present in both the training and test data but lower for groups absent from the training data (\cref{fig:fidelity}).
This indicates that the performance of the student is negatively influenced by covariate shift and does not accurately recover the teacher's decision boundary in these groups (see \cref{fig:teaser}). 
\cref{tab:disagreement} shows that the majority of samples where teacher and student disagree on the test set belongs to groups which are absent from the training data. Given a sample from the test set where teacher and student disagree, there is high empirical likelihood that this sample belongs to a group which is not present in the training data. 

\begin{figure}[tb]
    \centering
    \includegraphics[width=.99\columnwidth]{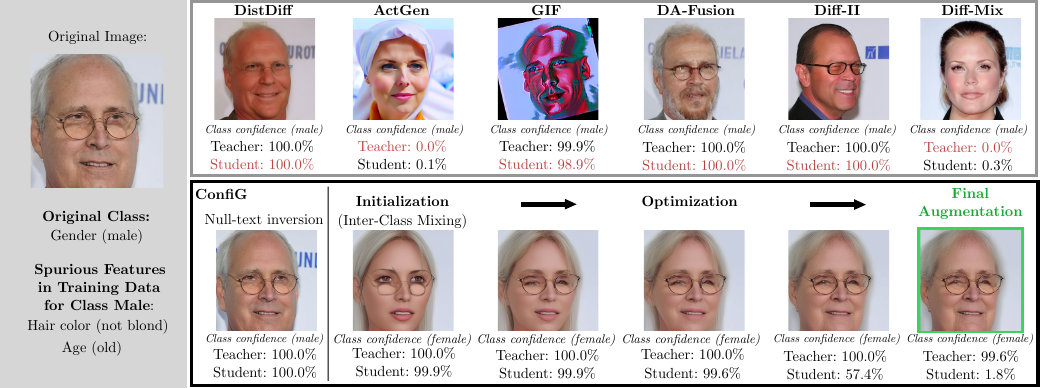}
    \caption{
    \textbf{Illustration of ConfiG and Other Diffusion-Based Augmentation Methods:}
    \textbf{Left:} 
    the task is gender prediction (male/female) where all men in the training data are old and not blond and all women in the training data are young and blond.
    \textbf{Right:} Existing augmentations yield high confidences for ``male'' or ``female'' for both the teacher and student. In contrast, ConfiG maximizes the difference between teacher and student and thus generates a challenging example for the student - in this case an old and not blond female face which has attributes that the student picked up as spurious features from the training data.
    }
    \label{fig:example_traj}
    \vskip-.5cm
\end{figure}
\noindent\textbf{Latent Optimization.} ConfiG aims to generate augmented images that reduce reliance on spurious features or dataset-specific biases without requiring explicit knowledge of what these features are.
Therefore, we make use of the observation that teacher and student predictions tend to diverge on samples from groups that are absent from the training data.
We use a diffusion model and real training images to generate augmented samples on which the student has low confidence in the target class while the teacher’s confidence remains high.
Maximizing the teacher’s confidence ensures that the causal, class-relevant features are preserved and the image label remains correct.
Simultaneously minimizing the student’s confidence yields images that include non-causal (potentially spurious) features which lead the student into making incorrect decisions.
For this purpose, we use the DiG-IN guided diffusion framework~\cite{dig_in,dash} based on Stable Diffusion~\cite{sd14}.
To simplify the notation below, we define $\mathcal{M}_{\epsilon_\theta}$ as the application of a noise prediction model $\epsilon_\theta$, iteratively mapping a latent variable $\textbf{z}_t$ to its final representation $\textbf{z}_T$.
For an image $\textbf{x}$ with corresponding original label $y$, we first encode the image $\textbf{z}_T = \mathcal{E}(\textbf{x})$ and apply null-text inversion~\cite{nulltext} to obtain a latent code $\textbf{z}_0$ and null-text $\emptyset$ suitable for reconstructing $\textbf{z}_T \approx \hat{\textbf{z}}_T = \mathcal{M}_{\epsilon_\theta}(\textbf{z}_0, \emptyset, C)$ where $C$ is a conditioning signal from a CLIP text encoding.
We further need to decode the latent code back to images $\textbf{x} \approx \hat{\textbf{x}} = \mathcal{D}(\hat{\textbf{z}}_T)$, as input for the teacher $\textbf{t}$ and student $\textbf{f}$.
We then minimize the loss:
\begin{align}
\underset{\textbf{z}_0, \emptyset, C}{\textrm{min}} \; \textbf{f}(\mathcal{D}\big(\mathcal{M}_{\epsilon_\theta}(\textbf{z}_0, \emptyset, C))\big)_y^\gamma + \Big(1 - \textbf{t}\big(\mathcal{D}(\mathcal{M}_{\epsilon_\theta}(\textbf{z}_0, \emptyset, C))\big)_y\Big)^\gamma
 \label{eq:loss}
\end{align}
As Problem~\eqref{eq:loss} is differentiable with respect to $\textbf{z}_0$, the null-text $\emptyset$ and the conditioning $C$, we can solve it using a stochastic optimizer such as Adam~\cite{adam}.
In practice, we observe that the student often is highly confident on the initial image with small gradients~\cite{dig_in}.
We alleviate this issue by adding the nonlinear transformation $(\cdot)^\gamma$ to improve optimization speed.
Empirically, we find $\gamma = 2$ to perform well. 
In addition, we follow DiG-IN and use a foreground-aware distance regularization.
An example of the optimization process is visualized in \cref{fig:example_traj}.
The closest baseline is ActGen~\cite{actgen}, which can generate augmentations that minimize the cross-entropy loss of the auxiliary student model given a real image and a target class.
However, it lacks explicit class guidance apart from the text prompt and the real image.
As we show in \cref{fig:example_traj}, ActGen may thus modify the causal attribute (i.e., gender) rather than only the spurious features.

\begin{wrapfigure}{r}{0.5\textwidth}
    \vskip-.6cm
    \centering
    \includegraphics[width=.49\textwidth]{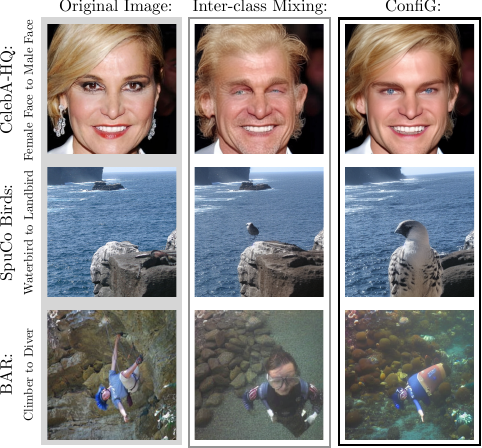}
    \caption{Images from CelebA-HQ, SpuCo Birds, and BAR. Example augmentations show inter-class mixing alone versus the full ConfiG pipeline, which combines this mixing with latent optimization. }
    \label{fig:example_celeba}
    \vskip-.7cm
\end{wrapfigure}

\noindent\textbf{Warm-Starting the Optimization through Inter-Class Mixing.} Diff-Mix~\cite{Diff-Mix} has demonstrated that inter-class mixing of training samples greatly benefits the effectiveness of diffusion-based data augmentations. For ConfiG, we combine inter-class mixing with latent optimization.
For every augmentation we sample a random target class, which is constrained to be \textit{different} from the original class.  
We warm-start the optimization process by first performing null-text inversion using the original image label, followed by a version of prompt-to-prompt  editing~\cite{p2p} modified for DiG-IN using the target class label.
The latent of the resulting image is taken as the initialization for the optimization problem of Equation~\eqref{eq:loss} which is optimized with respect to the target class. In \cref{fig:example_celeba} we visualize three example augmentations after only inter-class mixing and of ConfiG (with additional subsequent latent optimization).\\
\noindent \textbf{Auxiliary student.}
To generate images with ConfiG, we require a student network to optimize Equation~\eqref{eq:loss}.
We fine-tune an auxiliary student on only the real training data, where it picks up the spurious features and biases as demonstrated in \cref{fig:fidelity}.
It is then used to generate the synthetic augmentations, which replace real samples with probability $\alpha=0.5$ in the training of more robust students.
\\
\noindent \textbf{Theoretical Investigation,} We provide a theoretical motivation of the effect of using ConfiG augmentations in App.~\ref{sec:theoretical}.

\section{Experiments}\label{sec:experiments}
In our experiments, we empirically demonstrate the effectiveness of ConfiG to improve knowledge distillation under covariate shift. We first outline the experimental setup (Sec.~\ref{sec:experimental_setup}) before discussing the results on datasets with group shift (Sec.~\ref{sec:group_shift_datasets}), Spurious ImageNet (Sec.~\ref{sec:imagenet100}) and the ablations (Sec.~\ref{sec:ablations}).
\subsection{Experimental setup}\label{sec:experimental_setup}
\textbf{Datasets.} Our experiments are based on four datasets for image classification.
We use CelebA-HQ~\cite{celebahq,liu2015faceattributes}, SpuCo Birds~\cite{joshi2025challengesopportunitiesimprovingworstgroup} and BAR~\cite{lff} to investigate group shift as an example for covariate shift.
We subsample CelebA-HQ and SpuCo Birds to obtain a training dataset with 250 samples per class, \textit{fully} discarding certain groups from the training data.
For \textit{CelebA-HQ} we only retain training images of young, blond female celebrities as well as old, non-blond male celebrities for training.
The descriptive feature is the gender, other features are spuriously correlated only in the training split.
The test set spans all combinations of age (old/young), hair color (blond/non-blond) and gender (female/male) which leads to eight test groups.
\textit{SpuCo Birds} includes waterbirds or landbirds over water or land backgrounds, with bird type as the descriptive feature.
For SpuCo Birds, we only retain training images of waterbirds with water background and landbirds with land background such that the background becomes a binary spurious feature.
The test split also contains waterbirds with land background and vice versa.
The \textit{Biased Activity Recognition} dataset (BAR) contains images of six sport activities (climbing, diving, fishing, pole vaulting, motorsports racing, throwing) where the training images are biased towards distinct places per class. 
We sample 100 images per class for training.
The test images feature the same sport activities but were taken at different places which have no overlap with the places for the training set.
Our fourth dataset is a subset of \textit{ImageNet}~\cite{imagenet}.
We subsample a challenging training dataset with 100 training images for each of the 100 classes contained in Spurious ImageNet~\cite{spurious_imagenet}. 
We denote this setting by \textit{ImageNet-100}. 
Further details on training and test data selection can be found in App.~\ref{app:hyperparameter}.
\noindent\textbf{Diffusion-Based Data Augmentation.}
The diffusion-based augmentations are combined with the real training sets.
Following DA-Fusion, Diff-Mix and Diff-II, we replace a natural image by a synthetic augmentation with probability $\alpha=0.5$.
The hyperparameters of ConfiG can be found in App.~\ref{app:hyperparameter}. 
For ConfiG we generate one synthetic augmentation per image.
To ensure a fair comparison with the baselines, we adjust the number of synthetic augmentations generated by each method such that the computational cost of data generation, measured in FLOPs, is approximately matched. More details about the computational cost of ConfiG and the multipliers used for all baselines and datasets can be found in App.~\ref{app:cost}.
Experiments with more ConfiG augmentations per image can be found in Sec.~\ref{sec:ablations} and App.~\ref{app:more_auf}.
\\
\noindent\textbf{Baselines.} We compare ConfiG to six state-of-the-art diffusion-based data augmentation methods:
GIF~\cite{gid}, DistDiff~\cite{distdiff}, DA-Fusion~\cite{dafusion}, Diff-Mix~\cite{Diff-Mix}, Diff-II~\cite{diff-ii} and ActGen~\cite{actgen}.
All methods generate augmentations by applying a diffusion model to augment real images.
To ensure a fair comparison, we follow the baselines~\cite{gid,dafusion,distdiff} and use Stable Diffusion~1.4~\cite{sd14} as implicit teacher for all methods. 
In App.~\ref{sec:hyper_sd} we demonstrate that ConfiG also works with modern single-step diffusion models~\cite{hyper-sd} for faster generation.
\\
ActGen and ConfiG use a student model for generating data augmentations. For this purpose, we use the setup introduced in Sec.~\ref{sec:config}  and first distill an auxiliary student (which can pick up spurious features due to the limited coverage of the training data)
and then use this student to generate augmentations with the goal of getting close to the teacher model even outside the domain covered by the training data. In this way, the obtained student is more robust against picking up spurious features as shown by a significantly improved worst group accuracy.
Further details on the baselines are listed in App.~\ref{app:baseline summary}.\\
\noindent\textbf{Models and Training Hyperparameters.}
For our main experiments we use a ViT-T~\cite{vit} student pre-trained on ImageNet-21k using the AugReg recipe~\cite{augreg}.
Ablations with different student and teacher models are in App.~\ref{app:different_student_models}.
We use a CLIP~\cite{clip,open_clip} ViT-L/14 pre-trained on DataComp-XL~\cite{datacomp} as teacher. 
For CelebA-HQ and BAR we directly use classnames to obtain zero-shot predictions from the teacher. For SpuCo Birds we first use fine-grained bird categories and then aggregate into waterbirds and landbirds to obtain robust predictions (see App.~\ref{app:hyperparameter} for details).
The standard, non-synthetic augmentations used for training on CelebA-HQ, SpuCo Birds and BAR are random resized crops by default. Ablations with stronger, non-synthetic augmentations can be found in Sec.~\ref{sec:ablations}.
On ImageNet-100, we follow the setup of VanillaKD~\cite{VanillaKD} and use a BEiTv2-B~\cite{beitv2} teacher trained on the full ImageNet dataset together with the same ``A1'' augmentations~\cite{a1_aug} for our experiments. 
A detailed overview of all hyperparameters for training and standard augmentations can be found in App.~\ref{app:hyperparameter}.
\\
\noindent\textbf{Evaluation Metrics.} 
We consider three evaluation metrics for SpuCo Birds and CelebA-HQ. 
Given a test set $\mathcal{D}_{\mathrm{test}}=\{(\textbf{x}_i,y_i)\}_{i=1}^N$ composed of $K$ disjoint groups $\{\mathcal{D}_{\mathrm{test}}^{1},...,\mathcal{D}_{\mathrm{test}}^{K}\}$,
the sample mean accuracy (\textbf{SMA}) is the average accuracy independent of groups:
\begin{equation}
    \vspace{-.18cm}
    \textrm{\textbf{SMA}} (f) =\frac{1}{|\mathcal{D}_{\mathrm{test}}|} \sum_{(\textbf{x}_j,y_j)\in \mathcal{D}_{\mathrm{test}}}  \scalebox{1.2}{$\mathbbm{1}$}(\textrm{argmax}\;\textbf{f}(\textbf{x}_j)= y_j)
    \vspace{-.02cm}
\end{equation}
while the group mean accuracy (\textbf{GMA}) computes the average accuracy when groups are weighted equally:
\pagebreak
\begin{equation}
    \textrm{\textbf{GMA}} (f) = \frac{1}{K} \sum_{i=1}^K \frac{1}{|\mathcal{D}_{\mathrm{test}}^{i}|} \sum_{(\textbf{x}_j,y_j)\in \mathcal{D}_{\mathrm{test}}^{i}}  \scalebox{1.2}{$\mathbbm{1}$}(\textrm{argmax}\;\textbf{f}(\textbf{x}_j)= y_j).
\end{equation}
The worst group accuracy (\textbf{WGA}) is the minimum accuracy on a group:
\begin{equation}
    \textrm{\textbf{WGA}} (f) = \underset{
    i=1,\ldots,K
    } {\textrm{min}} \Big( \frac{1}{|\mathcal{D}_{\mathrm{test}}^{i}|} \sum_{(\textbf{x}_j,y_j)\in \mathcal{D}_{\mathrm{test}}^{i}} \scalebox{1.2}{$\mathbbm{1}$}(\textrm{argmax}\;\textbf{f}(\textbf{x}_j)= y_j) \Big).
\end{equation}
On SpuCo Birds, the test groups are balanced so group and sample mean accuracies coincide.
On BAR, the test data contains only one group per class which is disjoint from the groups in the training data such that we only evaluate the sample mean accuracy.
On Spurious ImageNet, we consider two evaluation metrics. First, the validation accuracy on all images from the corresponding classes of the ImageNet validation split. Second, the mean spurious score introduced by~\cite{spurious_imagenet}, computed using the predicted probability for the corresponding class, which we denote by SpuScore. 
The SpuScore measures the class-wise separation of images from Spurious ImageNet~\cite{spurious_imagenet}, containing the spurious feature but not the class object, versus the validation set of images of that class. The SpuScore averages these scores over all 100 classes.
\begin{table*}[bp!]
\centering
\begin{threeparttable}
\caption{
\textbf{Knowledge distillation using diffusion-based data augmentation under covariate shift:}
 We report mean in $\%$ and \textcolor{gray}{standard deviation} over six training runs. The student is a ViT-T distilled from a CLIP ViT-L teacher using EDRM with random resized crops as standard augmentations. Real Data refers to the student obtained using only real training images whereas the rows below correspond to the student models of different diffusion-based data augmentation methods. The highest score in each column is highlighted in \textbf{bold}, the second highest is \underline{underlined}. \vspace{-.2cm}}
\label{tab:group_shift}
\begin{tabular}{@{}lcccccc@{}}
\toprule
& \multicolumn{3}{c}{\textbf{CelebA-HQ}} & \multicolumn{2}{c}{\textbf{SpuCo Birds}} & \textbf{BAR} \\
\cmidrule(lr){2-4} \cmidrule(lr){5-6} \cmidrule(lr){7-7}
\textbf{Method} & \textbf{SMA} & \textbf{GMA} & \textbf{WGA} & \textbf{SMA} & \textbf{WGA} & \textbf{SMA} \\
\midrule
Teacher & 99.03 & 97.72 & 91.89 & 96.90 & 93.80 & 95.72 \\
\midrule
Real Data & 93.98\,\stdv{1.68} & 86.89\,\stdv{2.48} & 53.44\,\stdv{8.12} & 56.53\,\stdv{2.40} & 12.97\,\stdv{6.65} & 37.69\,\stdv{3.58} \\
\midrule
GIF & 96.70\,\stdv{0.87} & 90.78\,\stdv{1.00} & 68.92\,\stdv{7.60} & 63.15\,\stdv{1.20} & 25.90\,\stdv{2.53} & 47.91\,\stdv{3.30} \\
DistDiff & 96.33\,\stdv{0.58} & 90.41\,\stdv{1.11} & 70.25\,\stdv{5.46} & 63.94\,\stdv{2.33} & 24.07\,\stdv{4.62} & 45.91\,\stdv{3.36} \\
DA-Fusion & 95.39\,\stdv{1.52} & 90.09\,\stdv{0.78} & 67.96\,\stdv{6.79} & \underline{71.45}\,\stdv{1.54} & 30.20\,\stdv{5.18} & 53.03\,\stdv{3.34} \\
Diff-Mix &  \underline{97.36}\,\stdv{0.16} & \underline{93.07}\,\stdv{0.73} & 70.63\,\stdv{3.76} & 69.49\,\stdv{2.33} & \underline{32.13}\,\stdv{2.22} & \textbf{59.28}\,\stdv{2.32} \\
Diff-II & 95.08\,\stdv{1.75} & 89.48\,\stdv{1.41} & 64.96\,\stdv{7.62} & 61.88\,\stdv{1.80} & 22.90\,\stdv{4.25} & 56.40\,\stdv{2.34} \\
ActGen & 97.04\,\stdv{0.48} & 92.61\,\stdv{0.67} & \underline{75.68}\,\stdv{4.22} & 64.54\,\stdv{2.07} & 25.97\,\stdv{1.80} & 55.71\,\stdv{2.70} \\
ConfiG & \textbf{97.76}\,\stdv{0.25} & \textbf{95.67}\,\stdv{0.55} & \textbf{88.15}\,\stdv{2.10} & \textbf{73.38}\,\stdv{1.28} & \textbf{39.50}\,\stdv{4.53} & \underline{58.66}\,\stdv{3.76} \\
\bottomrule
\end{tabular}
\end{threeparttable}
\end{table*}
\subsection{Group Shift Datasets}\label{sec:group_shift_datasets}
\begin{figure}[tb]
    \centering
    \hspace{-.5cm}
    \includegraphics[width=.97\columnwidth]{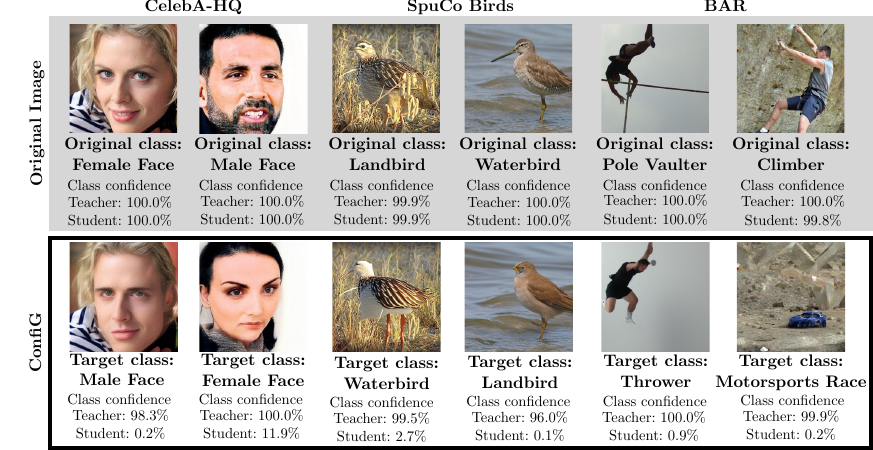}
    \vskip-0.1cm
    \caption{\textbf{Qualitative Examples of ConfiG} on CelebA-HQ, SpuCo Birds and BAR. Note that the target class is chosen randomly to be different from the original class. The images generated by ConfiG have low confidence of the student and show that it relies on spurious features: a young male with blond hair is recognized as female (CelebA-HQ) or a landbird standing in water as waterbird.
    }
    \label{fig:examples}
\end{figure}

\textbf{Quantitative Results.} \cref{tab:group_shift} summarizes results on CelebA-HQ, SpuCo Birds and BAR. We make two key observations: First, synthetic data augmentations generally improve the test performance in comparison to only training with real images. Second, ConfiG consistently yields the best or second-best SMA and best GMA and WGA results. Across all metrics, our method is at least on par with the best baseline despite using fewer synthetic augmentations.
On CelebA-HQ, the performance of the student using ConfiG augmentations is generally the highest, with a 1$\%$ gap to the teacher in SMA and a 4$\%$ gap in WGA. In comparison to only real data, the absolute WGA improved by over 34$\%$.
On the SpuCo Birds and BAR datasets, a larger gap to the teacher remains of over 40$\%$ and 58$\%$ in SMA when training with real data only. ConfiG augmentations reduce this difference to 23$\%$ and 37$\%$ respectively.
We demonstrate that combining stronger, non-synthetic data augmentations with ConfiG can further improve the student performance in Sec.~\ref{sec:ablations}.\\
\noindent\textbf{Qualitative Results.} We show six examples for ConfiG augmentations in \cref{fig:examples}.
The ConfiG augmentations retain high teacher confidence while reducing the student confidence.
Further examples can be found in App.~\ref{app:further_examples}.

\subsection{ImageNet-100}\label{sec:imagenet100}
\cref{tab:in100} presents results for fine-tuning ViT-T and ViT-S students on ImageNet-100.
Synthetic data augmentations generally improve the validation further, with ConfiG achieving the best validation accuracies and spurious scores. 
For the ViT-T student, ConfiG improves the validation accuracy by more than 1.6$\%$ and the spurious score by more than 1.8$\%$ in comparison to all baselines.
The improvements for the ViT-S student by 0.6$\%$ in validation accuracy and by 0.9$\%$ in spurious score are
\begin{wrapfigure}{r}{0.6\textwidth}
\vspace{-.4cm}
\centering
\begin{threeparttable}
\captionof{table}{\textbf{Results on ImageNet-100.} ConfiG improves the SpuScore in comparison to all baselines and yields the best validation accuracy. The reported values are in $\%$.  The highest score in each column is highlighted in \textbf{bold}, the second highest is \underline{underlined}. The teacher has a SMA of 96.12$\%$ and a SpuScore of 85.45. \vspace{.4cm}}
\label{tab:in100}
\begin{tabular}{@{}lcccc@{}}
\toprule
& \multicolumn{2}{c}{\textbf{ViT-T}} & \multicolumn{2}{c}{\textbf{ViT-S}} \\
\cmidrule(lr){2-3} \cmidrule(lr){4-5}
\textbf{Method} & \textbf{SMA} & \textbf{SpuScore} & \textbf{SMA} & \textbf{SpuScore} \\
\midrule
Real data & 71.46 & 54.05 & 86.46 & 68.62 \\
\midrule
GIF & 76.22 & 58.55 & 89.36 & 72.64 \\
DistDiff & 76.84 & 59.13 & 90.66 & \underline{74.59} \\
DA-Fusion & 74.92 & 56.99 & 89.84 & 73.40 \\
Diff-Mix &  76.88 & 59.48 & 89.98 & 72.32 \\
Diff-II & 74.76 & 58.43 & 89.00 & 74.53 \\
ActGen & \underline{78.36} & \underline{60.48} & \underline{91.04} & 73.17 \\
ConfiG & \textbf{80.02} & \textbf{62.36} & \textbf{91.68} & \textbf{75.49} \\
\bottomrule
\end{tabular}
\end{threeparttable}
\vskip-1.6cm
\end{wrapfigure}
slightly smaller. These results show that ConfiG consistently enhances knowledge distillation under covariate shift, not only in group-based settings but also for class-specific spurious features.

\subsection{Ablations}
\label{sec:ablations}
We perform ablation studies on the group shift datasets to investigate the impact of inter-class mixing and latent optimization and analyze the effect of more synthetic and stronger non-synthetic augmentations. 
Furthermore, we investigate the influence of weaker teacher models and demonstrate that ConfiG improves the performance of model-centric bias mitigation methods.

\begin{table*}[bp!]
\centering
\begin{threeparttable}
\caption{\textbf{Only Inter-class Mixing or Latent Optimization compared to ConfiG.} Using only latent optimization after null-text inversion without inter-class mixing (using the original class as target class) or purely inter-class mixing with prompt-to-prompt (p2p) editing as data augmentation improves the performance in comparison to training on real data only. Combining both into ConfiG achieves the best results.}
\label{tab:components}
\begin{tabular}{@{}lcccccc@{}}
\toprule
& \multicolumn{3}{c}{\textbf{CelebA-HQ}} & \multicolumn{2}{c}{\textbf{SpuCo Birds}} & \textbf{BAR} \\
\cmidrule(lr){2-4} \cmidrule(lr){5-6} \cmidrule(lr){7-7}
\textbf{Method} & \textbf{SMA} & \textbf{GMA} & \textbf{WGA} & \textbf{SMA} & \textbf{WGA} & \textbf{SMA} \\
\midrule
Real Data & 93.98\stdv{1.68} & 86.89\stdv{2.48} & 53.44\stdv{8.12} & 56.53\stdv{2.40} & 12.97\stdv{6.65} & 37.69\stdv{3.58} \\
Latent Opt. & 93.35\stdv{1.33} & 89.01\stdv{1.13} & 71.11\stdv{3.74} & 61.53\stdv{2.03} & 21.63\stdv{3.61} & 39.40\stdv{3.53} \\
Only p2p & 97.40\stdv{0.23} & 94.74\stdv{0.55} & 86.99\stdv{1.61} & 66.64\stdv{1.83} & 33.20\stdv{3.41} & 44.09\stdv{3.77} \\
ConfiG & \textbf{97.76}\,\stdv{0.25} & \textbf{95.67}\,\stdv{0.55} & \textbf{88.15}\,\stdv{2.10} & \textbf{73.38}\,\stdv{1.28} & \textbf{39.50}\,\stdv{4.53} & \textbf{58.66}\,\stdv{3.76} \\
\bottomrule
\end{tabular}
\end{threeparttable}
\end{table*}
\noindent\textbf{Individual Components of ConfiG.} As introduced in Sec.~\ref{sec:config}, ConfiG combines inter-class mixing through prompt-to-prompt editing with latent optimization.
We assess the effect of these components individually in \cref{tab:components}. Here only latent optimization refers to optimizing Problem \eqref{eq:loss} without inter-class mixing which means the target class is the same original class.
While both components individually improve performance over training on purely real data, inter-class mixing alone is sufficient to be on par or even outperform the strongest baselines on CelebA-HQ. On BAR only performing inter-class mixing through prompt-to-prompt editing is less effective.
Latent optimization directly after null-text inversion performs worse than only inter-class mixing.
However, one gets a substantial boost in performance in all evaluation metrics when combining both techniques as done in ConfiG.\\
\noindent\textbf{Stronger Standard Augmentations.} We report the results of distilling students with stronger, non-synthetic augmentations in \cref{tab:stronger_aug_main}. We compare ConfiG to Diff-Mix as the strongest baseline. 
The complete results for all baselines can be found in App.~\ref{app:stronger_aug_full}.
Compared to \cref{tab:group_shift}, we find that on BAR stronger standard augmentations with AutoAugment~\cite{aa} and MixUp~\cite{mixup} considerably improve the student performance, outperforming synthetic data augmentations with only random resized crops.
On CelebA-HQ and SpuCo Birds all standard augmentations with only real data remain worse in GMA and WGA than ConfiG with purely random resized crops. In all cases, the best overall performance can be achieved when combining stronger standard augmentations with ConfiG.
\begin{table*}[t]
\centering
\begin{threeparttable}
\caption{\textbf{Stronger Standard Data Augmentation.} Combining stronger, non-synthetic augmentations with diffusion-based augmentations further improves the student performance both for Diff-Mix and ConfiG. ConfiG yields stronger gains than Diff-Mix. \vspace{-.15cm}}
\label{tab:stronger_aug_main}
\vskip-.3cm
\begin{tabular}{@{}lcccccc@{}}
\toprule
& \multicolumn{3}{c}{\textbf{CelebA-HQ}} & \multicolumn{2}{c}{\textbf{SpuCo Birds}} & \textbf{BAR} \\
\cmidrule(lr){2-4} \cmidrule(lr){5-6} \cmidrule(lr){7-7}
\textbf{Method} & \textbf{SMA} & \textbf{GMA} & \textbf{WGA} & \textbf{SMA} & \textbf{WGA} & \textbf{SMA} \\
\midrule
\multicolumn{7}{c}{\itshape ~~~Stronger Random Resized Crops + Flips} \\
Real Data & 97.17\,\stdv{0.66} & 91.81\,\stdv{0.77} & 69.97\,\stdv{7.19} & 60.58\,\stdv{0.79} & 17.63\,\stdv{2.77} & 52.17\,\stdv{3.98} \\
Diff-Mix & 97.69\,\stdv{0.38} & 93.58\,\stdv{1.02} & 72.62\,\stdv{6.69} & 70.28\,\stdv{3.28} & 33.40\,\stdv{6.39} & 64.91\,\stdv{1.16} \\
ConfiG & \textbf{98.42}\,\stdv{0.05} & \textbf{96.72}\,\stdv{0.30} & \textbf{90.98}\,\stdv{1.40} & \textbf{73.78}\,\stdv{1.60} & \textbf{42.10}\,\stdv{5.00} & \textbf{65.39}\,\stdv{1.91} \\
\midrule
\multicolumn{7}{c}{\itshape ~~~Stronger Random Resized Crops + Flips + AutoAugment~\cite{aa}} \\
Real Data & 97.82\,\stdv{0.11} & 92.50\,\stdv{0.67} & 72.22\,\stdv{3.76} & 65.46\,\stdv{0.80} & 25.17\,\stdv{1.99} & 61.65\,\stdv{2.99} \\
Diff-Mix & 97.86\,\stdv{0.21} & 93.60\,\stdv{0.81} & 74.21\,\stdv{5.61} & 72.63\,\stdv{2.24} & \underline{36.50}\,\stdv{7.86} & 67.89\,\stdv{2.27} \\
ConfiG & \textbf{98.53}\,\stdv{0.05} & \textbf{96.55}\,\stdv{0.36} & \textbf{89.45}\,\stdv{1.59} & \textbf{78.62}\,\stdv{1.61} & \textbf{50.53}\,\stdv{5.34} & \textbf{69.78}\,\stdv{0.70} \\
\midrule
\multicolumn{7}{c}{\itshape ~~~Stronger Random Resized Crops + Flips + MixUp~\cite{mixup}} \\
Real Data & 97.66\,\stdv{0.43} & 93.15\,\stdv{1.08} & 74.35\,\stdv{8.53} & 68.99\,\stdv{0.71} & 37.20\,\stdv{1.31} & 61.70\,\stdv{1.71} \\
Diff-Mix & \underline{98.06}\,\stdv{0.08} & 94.38\,\stdv{0.60} & 77.65\,\stdv{3.60} & 73.19\,\stdv{1.28} & \underline{43.00}\,\stdv{4.93} & 65.62\,\stdv{1.10} \\
ConfiG & \textbf{98.42}\,\stdv{0.11} & \textbf{96.79}\,\stdv{0.48} & \textbf{89.95}\,\stdv{1.38} & \textbf{77.99}\,\stdv{1.97} & \textbf{51.57}\,\stdv{5.73} & \textbf{68.81}\,\stdv{1.86} \\
\midrule
\multicolumn{7}{c}{\itshape ~~~Stronger Random Resized Crops + Flips + CutMix~\cite{cutmix}} \\
Real Data & 97.74\,\stdv{0.26} & 93.17\,\stdv{0.77} & 70.24\,\stdv{4.59} & 70.05\,\stdv{2.15} & 36.63\,\stdv{4.29} & 58.21\,\stdv{2.61} \\
Diff-Mix & 97.77\,\stdv{0.07} & 93.50\,\stdv{0.40} & 71.43\,\stdv{2.41} & 71.48\,\stdv{2.37} & \underline{38.97}\,\stdv{3.55} & 62.31\,\stdv{1.69} \\
ConfiG & \textbf{98.33}\,\stdv{0.21} & \textbf{96.14}\,\stdv{0.67} & \textbf{87.90}\,\stdv{2.80} & \textbf{77.66}\,\stdv{0.52} & \textbf{53.10}\,\stdv{3.42} & \textbf{65.34}\,\stdv{1.75} \\
\bottomrule
\end{tabular}
\vspace{-.5cm}
\end{threeparttable}
\end{table*}
\begin{table*}[tp!]
    \centering
    \caption{
    \textbf{Using two augmentations per image for ConfiG.} 
    The results with Diff-Mix are obtained with 56 augmentations per image to approximately match the generation costs in FLOPs (see App.~\ref{app:cost}).
    The replacement probability $\alpha$ is kept at $0.5$, $\delta$ refers to the difference to the results from \cref{tab:group_shift}. 
    }
    \label{tab:more_augs}
    \vskip-.02cm
    \begin{tabular}{@{}lcccccc@{}}
    \toprule
      & \multicolumn{3}{c}{\textbf{CelebA-HQ}} & \multicolumn{2}{c}{\textbf{SpuCo Birds}} & \textbf{BAR} \\
    \cmidrule(lr){2-4} \cmidrule(lr){5-6} \cmidrule(lr){7-7}
    \textbf{Method} & \textbf{SMA} & \textbf{GMA} & \textbf{WGA} & \textbf{SMA} & \textbf{WGA} & \textbf{SMA} \\
    \midrule
    Diff-Mix & 97.61\,\stdv{0.19} & 93.62\,\stdv{0.44} & 72.62\,\stdv{2.83} & 69.85\,\stdv{2.01} & 34.53\,\stdv{5.30} & 61.44\,\stdv{2.11} \\
     $\delta_{28\rightarrow 56}$ & \textcolor{darkgray}{\footnotesize{+0.25}} & \textcolor{darkgray}{\footnotesize{+0.55}} & \textcolor{darkgray}{\footnotesize{+1.99}} & \textcolor{darkgray}{\footnotesize{+0.36}} & \textcolor{darkgray}{\footnotesize{+2.40}} & \textcolor{darkgray}{\footnotesize{+2.16}} \\
    \midrule
    ConfiG & \textbf{97.75}\,\stdv{0.18} & \textbf{95.67}\,\stdv{0.34} &
    \textbf{87.83}\,\stdv{1.64} &
    \textbf{74.64}\stdv{1.72} & \textbf{46.90}\stdv{6.36} & \textbf{63.12}\,\stdv{1.71} \\
    
     $\delta_{1\rightarrow 2}$ & \textcolor{darkgray}{\footnotesize{-0.01}} & \textcolor{darkgray}{\footnotesize{+0.00}} & \textcolor{darkgray}{\footnotesize{-0.32}} & \textcolor{darkgray}{\footnotesize{+1.26}} & \textcolor{darkgray}{\footnotesize{+7.40}} & \textcolor{darkgray}{\footnotesize{+4.46}} \\
     
    \bottomrule
    \end{tabular}
    \vskip-0.2cm
\end{table*}
\noindent \textbf{Larger Number of Synthetic Augmentations.} \cref{tab:more_augs} shows results with two augmentations per image for ConfiG and 56 instead of 28 augmentations per image for Diff-Mix as the strongest baseline matched by the approximate generation FLOPs. Results for the remaining baselines can be found in App.~\ref{app:more_auf}. On CelebA-HQ the student performance remains close to the main results from \cref{tab:group_shift} for both ConfiG and Diff-Mix. On SpuCo Birds and BAR the performance of ConfiG improves with twice as many augmentations and the gap between Diff-Mix and ConfiG increases. This indicates more saturated performances with\\
\begin{wrapfigure}{r}{0.72\textwidth}
\vskip-.4cm
\centering
\begin{threeparttable}
\captionof{table}{\textbf{Results with Weaker Teacher Models.} We replace the CLIP ViT-L/14 used in Sec.~\ref{sec:group_shift_datasets} by weaker teachers (RN50 for CelebA-HQ and CLIP ViT-B/32 for BAR) which decreases the student performance but ConfiG still improves the performance over only real data and the Diff-Mix baseline. \vspace{.05cm}}
\label{tab:weaker_teacher_tab_main}

\begin{tabular}{@{}lcccc@{}}
\toprule
& \multicolumn{3}{c}{\textbf{CelebA-HQ}} & \multicolumn{1}{c}{\textbf{BAR}} \\
\cmidrule(lr){2-4} \cmidrule(lr){5-5}
\textbf{Method} & \textbf{SMA} & \textbf{GMA} & \textbf{WGA} & \textbf{SMA}  \\
\midrule
Teacher & 96.48 & 92.23 & 75.68 & 59.63 \\
\midrule
Real Data &
94.37\stdv{0.47} & 83.37\stdv{1.48} & 38.36\stdv{5.37} &
36.65\stdv{1.68}  \\
\midrule
Diff-Mix &
94.48\,\stdv{0.73} & 86.62\,\stdv{1.33} & 50.79\,\stdv{4.99} &
40.67\,\stdv{2.82}  \\
ConfiG &
\textbf{95.25}\stdv{0.52} & \textbf{90.16}\stdv{0.63} & \textbf{67.72}\stdv{2.91} &
\textbf{44.55}\stdv{1.76}  \\
\bottomrule
\end{tabular}
\vskip-.45cm
\end{threeparttable}
\end{wrapfigure}
Diff-Mix while ConfiG demonstrates better scaling.\\
\noindent\textbf{Weaker Teachers.} Since the teacher influences both the generated augmentations and the downstream training of the student model when using ConfiG, we investigate the effect of using weaker teacher models. We exchange the CLIP ViT-L/14 used in Sec.~\ref{sec:group_shift_datasets} by a CLIP RN50 trained on YFCC15M for the CelebA-HQ dataset and a CLIP ViT-B/32 trained on DataComp M for BAR. The student model and all hyperparameters remain the same as in Sec.~\ref{sec:group_shift_datasets}. \cref{tab:weaker_teacher_tab_main} demonstrates that a weaker teacher can indeed decrease the student performance. However, ConfiG still improves the performance over real data and Diff-Mix. The complete results for all baselines can be found in App.~\ref{app:weaker_teacher}.\\
\noindent\textbf{Data-Centric vs. Model-Centric Bias Mitigation.} The goal of ConfiG is to use synthetic data augmentations to mitigate the problem of limited coverage of the training data which can lead to biases and shortcut learning,
particularly in scenarios where certain subgroups are entirely absent.
This approach complements existing strategies that focus on learning robust models from a fixed dataset with potential biases~\cite{effbias}.
We demonstrate this by exemplarily combining ConfiG with pure ERM, LfF~\cite{lff} and uLA~\cite{ula}, three approaches for training classification models on biased datasets without bias information.
Since LfF and uLA are based on ERM, we label the synthetic data augmentations using the teacher before training.
In \cref{tab:model_bias} we show that LfF, uLA, and pure ERM are substantially improved with synthetic data augmentations from ConfiG.

\begin{table*}[tbp]
\centering
\begin{threeparttable}
\caption{\textbf{Using ConfiG Augmentations Improves Model-Centric Methods for Bias Mitigation.} We compare pure ERM, LfF~\cite{lff} and uLA~\cite{ula} with and without ConfiG augmentations. We labeled the ConfiG augmentations using the teacher prediction.}
\vskip-.05cm
\label{tab:model_bias}
\begin{tabular}{@{}lcccccc@{}}
\toprule
& \multicolumn{3}{c}{\textbf{CelebA-HQ}} & \multicolumn{2}{c}{\textbf{SpuCo Birds}} & \textbf{BAR} \\
\cmidrule(lr){2-4} \cmidrule(lr){5-6} \cmidrule(lr){7-7}
\textbf{Method} & \textbf{SMA} & \textbf{GMA} & \textbf{WGA} & \textbf{SMA} & \textbf{WGA} & \textbf{SMA} \\
\midrule
ERM & 90.34\stdv{3.24} & 84.94\stdv{1.89} & 57.84\stdv{3.93} & 53.17\stdv{1.03} & 3.67\stdv{2.27} & 38.43\stdv{1.88} \\
uLA & 84.51\stdv{3.72} & 79.64\stdv{2.43} & 45.57\stdv{7.38} & 53.16\stdv{0.79} & 3.37\stdv{1.60} & 39.14\stdv{3.94} \\
LfF & 87.98\stdv{5.25} & 82.60\stdv{3.84} & 51.22\stdv{7.15} & 52.47\stdv{0.40} & 2.87\stdv{1.22} & 35.55\stdv{1.60} \\
\midrule
ERM{\tiny$^+$}ConfiG &
97.46\stdv{0.16} & 95.52\stdv{0.40} & 90.25\stdv{1.09} & 69.26\stdv{3.99} & 40.57\stdv{7.89} & \textbf{58.33}\stdv{2.42} \\
uLA{\tiny$^+$}ConfiG & 97.33\stdv{0.33} & 95.30\stdv{0.90} & 89.12\stdv{4.97} & \textbf{70.38}\stdv{2.72} & \textbf{40.87}\stdv{4.14} & 57.75\stdv{1.92} \\
LfF{\tiny$^+$}ConfiG & \textbf{97.49}\stdv{0.28} & \textbf{95.82}\stdv{0.47} & \textbf{90.72}\stdv{1.92} & 67.86\stdv{0.64} & 32.93\stdv{3.07} & 56.75\stdv{4.18} \\
\bottomrule
\end{tabular}
\end{threeparttable}
\vskip-.15cm
\end{table*}

\vspace{-.1cm}
\section{Conclusion}
\vspace{-.1cm}
We introduce ConfiG, a novel confidence-guided data augmentation method designed to enhance knowledge distillation under unknown covariate shift.
Our approach effectively addresses the challenge of group shift or class-level spurious features in training data by generating targeted augmentations that maximize the disagreement between a robust teacher and biased student models.
Experimental results on datasets such as CelebA-HQ, SpuCo Birds, and BAR demonstrate that ConfiG improves both worst group and mean group accuracy in a group shift setting and
is best or second-best in sample mean accuracy.
On ImageNet-100, ConfiG improves the validation accuracy and SpuScore, indicating increased robustness of the student against class-wise spurious features.

\noindent \textbf{Limitations and Future Work.}
We have demonstrated ConfiG in a classification setting where images are augmented globally. Location-aware tasks such as object detection require constraining the image augmentation process~\cite{object_det}, which we leave for future work.
Furthermore, our experimental setup requires pre-trained teacher models for knowledge distillation and data generation for both ConfiG and the baselines. 
As dataset and model sizes continue to grow, general purpose models such as CLIP and Stable Diffusion are suitable teachers for a wide range of domains.
Future research could explore dedicated combinations of synthetic data augmentation and model-centric bias mitigation strategies.
Finally, while we demonstrate that ConfiG works with Hyper-SD as a modern diffusion backbone in App.~\ref{sec:hyper_sd}, we believe that these initial results can be further improved.
We leave the exploration of other model types, e.g. context-guided generation~\cite{labs2025flux1kontextflowmatching,wu2025qwenimagetechnicalreport}, to future work.

\vspace{-.2cm}
\section*{Acknowledgements}
\vspace{-.2cm}
We would like to thank Maximilian Augustin and Yannic Neuhaus for support on the DiG-IN implementation as well as Dan Zhang for helpful discussions on the manuscript. We also thank the European Laboratory for
Learning and Intelligent Systems (ELLIS) for supporting Niclas Popp.

\newpage

\bibliographystyle{splncs04}
\bibliography{cleaned}

\setcounter{section}{0}
\appendix
\section{Theoretical Investigation}
\label{sec:theoretical}

In this section, we provide a theoretical motivation why ConfiG data augmentations can decrease $|\Omega|= |\mathcal{R}_{\text{test}}(\textbf{f}) - \mathcal{R}_{\text{train}}(\textbf{f})|$.
The notation is consistent with Sec.~\ref{sec:background}.
When augmenting the training dataset $\mathcal{D}_{train}$ with samples from $\mathcal{Q}$, which we define as the distribution of augmented data points, the resulting distribution can be written as a convex combination of $\mathcal{P}_{\mathrm{train}}$ and $\mathcal{Q}$
\begin{equation}\mathcal{P}_{\mathrm{aug}}=(1-\alpha)\,\mathcal{P}_{\mathrm{train}}+\alpha\,\mathcal{Q},\qquad
0<\alpha<1.
\end{equation}
First, we show that data augmentations that are more challenging for the student than for the teacher can reduce the student’s generalization error $|\Omega|$. Second, we motivate why ConfiG can generate such data augmentations.

\begin{restatable}{proposition}{mainprop}\label{prop:1}\textbf{(Decreasing $|\Omega|$ through confidence‐guided augmentations)}
Let $\ell$ be the cross-entropy loss and assume that under covariate shift the student has low training error but high test error
\begin{equation}
\Omega \;=\;\mathcal{R}_{\mathrm{test}}(\mathbf{f}) - \mathcal{R}_{\mathrm{train}}(\mathbf{f})>0.
\end{equation}
Let the teacher provide an accurate estimate of the true data-generating distribution $\textbf{p}^*$ in terms of the KL divergence
\begin{equation}\label{eq:teacherbound}
    KL\!\bigl(\textbf{p}^*(\textbf{x})\,\|\,\textbf{t}(\textbf{x})\bigr) \leq \kappa_t < \infty \; \text{for all} \; \textbf{x}\in\text{supp}(\mathcal{P}_{\mathrm{test}}) \cup \text{supp}(\mathcal{P}_{\mathrm{train}}) \cup \text{supp}(\mathcal{Q}).
\end{equation}
Assume that the data augmentations are more challenging for the student than for the teacher
\begin{equation}\label{eq:student_teacher_approx}
     \mathcal{R}_\mathcal{Q} (\textbf{f}) -\mathcal{R}_\mathcal{Q}(\textbf{t})\geq \epsilon_{aug}>0
\end{equation}
while the training data is similarly challenging for both student and teacher
\begin{equation}\label{eq:student_teacher_approx_2}
    0<\mathcal{R}_{\mathrm{train}}(\textbf{f})-\mathcal{R}_{\mathrm{train}}(\textbf{t}) \le \epsilon_{train}.
\end{equation}
Define $\Delta_{\textbf{p}}=\bigl|\mathcal{R}_{\mathrm{train}}(\textbf{p}^*) - \mathcal{R}_{\mathcal{Q}}(\textbf{p}^*)\bigr|$
and $\Delta_{\textbf{f}}=\bigl|\mathcal{R}_{\mathrm{train}}(\mathbf{f}) - \mathcal{R}_{\mathcal{Q}}(\mathbf{f})\bigr|$
. If $\Theta = \epsilon_{aug} -\bigl(\kappa_t + \Delta_{\textbf{p}}\bigr) - \epsilon_{train} >0$ and $0<\alpha\le \min\{1,\frac{3\Omega}{2\Delta_{\textbf{f}}}\}$ then the augmented generalization error
\begin{equation}
\Omega_{\mathrm{aug}}
=\mathcal{R}_{\mathrm{test}}(\mathbf{f})
-\bigl[(1-\alpha)\mathcal{R}_{\mathrm{train}}(\mathbf{f})+\alpha \mathcal{R}_Q(\mathbf{f})\bigr]
\end{equation}
satisfies
\begin{equation}
|\Omega_{\mathrm{aug}}|\leq \max\{\Omega-\alpha \Theta,\frac{\Omega}{2}\} < \Omega.
\end{equation}
\end{restatable}
We first prove an auxiliary lemma that bounds the teacher‐risk between train and augmentation distribution under the cross-entropy loss.
\begin{lemma} Let $\ell$ be the cross-entropy loss. If $\mathcal{Q}$ and $\textbf{t}$ satisfy the assumptions from Proposition~\ref{prop:1}, then
\begin{equation}
\bigl|\mathcal{R}_{\mathrm{train}}(\textbf{t}) - \mathcal{R}_{\mathcal{Q}}(\textbf{t})\bigr|
\;\le\;
\kappa_t \;+\; \Delta_{\textbf{p}}.
\end{equation}
\end{lemma}

\begin{proof}
First, we write the teacher-loss at $\textbf{x}$ in terms of the data-generating conditional $\textbf{p}^*(\textbf{x})$:
\begin{equation}
\mathbb{E}_{y|x}[\ell(\textbf{t}(\textbf{x}),y)]
= -\sum_y \textbf{p}^*_y(\textbf{x})\log \textbf{t}_y(\textbf{x})
= H\bigl(\textbf{p}^*(\textbf{x})\bigr) + KL\!\bigl(\textbf{p}^*(\textbf{x})\,\|\,\textbf{t}(\textbf{x})\bigr),
\end{equation}
where $H(\textbf{p})$ is the entropy of $\textbf{p}$. Taking expectations over $\textbf{x}$ from $\mathcal{P}_{\mathrm{train}}$ and $\mathcal{Q}$ respectively we get
\begin{align}
\mathcal{R}_{\mathrm{train}}(\textbf{t})-\mathcal{R}_{\mathcal{Q}}(\textbf{t})
& =
%\bigl[
\mathcal{R}_{\mathrm{train}}(\textbf{p}^*)-\mathcal{R}_{\mathcal{Q}}(\textbf{p}^*)
%\bigr] 
\\
 & +
\bigl[ \mathbb{E}_{x\sim\mathcal{P}_{\mathrm{train}}} KL(\textbf{p}^*(\textbf{x})\|\textbf{t}(\textbf{x}))
- \mathbb{E}_{x\sim\mathcal{Q}} KL(\textbf{p}^*(\textbf{x})\|\textbf{t}(\textbf{x})) \bigr]. \label{eq:KL-diff}
\end{align}
By definition $|\mathcal{R}_{\mathrm{train}}(\textbf{p}^*)-\mathcal{R}_{\mathcal{Q}}(\textbf{p}^*)|\le\Delta_{\textbf{p}}$.
Moreover, by Equation~\eqref{eq:teacherbound} each KL term is bounded by $\kappa_t$. Due to the positivity of the KL-divergence the second term in~\eqref{eq:KL-diff} is thus contained in the interval $[-\kappa_t,\kappa_t]$. Combining these two bounds yields the result.
\qed
\end{proof}

\noindent Next, we prove Proposition~\ref{prop:1}:

\begin{proof}
First, decompose for any distribution $D$,
\begin{equation}
\mathcal{R}_{(\textbf{x},y)\sim \mathcal{D}} (\textbf{f}) = \mathcal{R}_{(\textbf{x},y)\sim \mathcal{D}} (\textbf{t}) + \bigl(\mathcal{R}_{(\textbf{x},y)\sim \mathcal{D}}(\textbf{f}) - \mathcal{R}_{(\textbf{x},y)\sim \mathcal{D}}(\textbf{t}) \bigr).
\end{equation}
Thus, we can write the difference between the risk on $\mathcal{P}_{train}$ and $\mathcal{Q}$ as
\begin{align}
\mathcal{R}_\mathcal{Q}(\textbf{f}) - \mathcal{R}_{\mathrm{train}}(\textbf{f})
&=\bigl(\mathcal{R}_\mathcal{Q} (\textbf{t}) - \mathcal{R}_{\mathrm{train}}(\textbf{t})\bigr) \\
 &+ \bigl( \mathcal{R}_\mathcal{Q}(\textbf{f})
 -\mathcal{R}_\mathcal{Q}(\textbf{t})\bigr) \\
 &-(\mathcal{R}_{\mathrm{train}}(\textbf{f})-\mathcal{R}_{\mathrm{train}}(\textbf{t})).
\end{align}
By the Lemma we get
\begin{equation}\label{eq:e_1}
\mathcal{R}_\mathcal{Q}(\textbf{t}) - \mathcal{R}_{\mathrm{train}}(\textbf{t})  \ge -\bigl(\kappa_t+\Delta_{\textbf{p}}\bigr).
\end{equation}
From the assumptions we can directly bound the difference of the student and teacher risk on $\mathcal{Q}$
\begin{equation}\label{eq:e_2}
\mathcal{R}_\mathcal{Q} (\textbf{f}) -\mathcal{R}_\mathcal{Q}(\textbf{t}) \geq \epsilon_{aug}>0,
\end{equation}
and the difference in risk between teacher and student on $\mathcal{P}_{\mathrm{train}}$
\begin{equation}\label{eq:e_3}
\mathcal{R}_{\mathrm{train}}(\textbf{f})-\mathcal{R}_{\mathrm{train}}(\textbf{t}) \le \epsilon_{train}.
\end{equation}
By combining~\eqref{eq:e_1},~\eqref{eq:e_2} and~\eqref{eq:e_3}, we get
\begin{equation}
\mathcal{R}_\mathcal{Q}(\textbf{f}) - \mathcal{R}_{\mathrm{train}}(\textbf{f})
\ge  
\epsilon_{aug} -\bigl(\kappa_t + \Delta_{\textbf{p}}\bigr) - \epsilon_{train}
=\Theta > 0.
\end{equation}
Thus,
\begin{align}
\Omega_{\mathrm{aug}}
&=\mathcal{R}_{\mathrm{test}}(\textbf{f})
-\bigl((1-\alpha)\mathcal{R}_{\mathrm{train}}(\textbf{f})+\alpha \mathcal{R}_\mathcal{Q}(\textbf{f})\bigr)
\\
 &=\mathcal{R}_{\mathrm{test}}(\textbf{f}) - \mathcal{R}_{\mathrm{train}}(\textbf{f}) - \alpha\,\bigl(\mathcal{\mathcal{R}}_\mathcal{Q}(\textbf{f}) -\mathcal{\mathcal{R}}_{\mathrm{train}}(\textbf{f})\bigr)\\
 & =\Omega - \alpha\,\bigl(\mathcal{\mathcal{R}}_\mathcal{Q}(\textbf{f})-\mathcal{\mathcal{R}}_{\mathrm{train}}(\textbf{f})\bigr)  \\
 & \le \Omega - \alpha\,\Theta
\end{align}
In summary, we obtain
\begin{equation}
    \Omega_{aug} \in [\Omega - \alpha \Delta_{\textbf{f}}, \Omega - \alpha \Theta]
\end{equation}
Since $0<\alpha\le \min\{1,\frac{3\Omega}{2\Delta_{\textbf{f}}}\}$, we have
\begin{equation}
\bigl|\Omega_{\mathrm{aug}}\bigr|\leq \max\{\Omega-\alpha\Theta,\frac{\Omega}{2}\}.
\end{equation}
\qed
\end{proof}

\noindent To underline the practical relevance of Proposition~\ref{prop:1}, we motivate why ConfiG can generate data augmentations which are challenging for the student in the sense of Equation~\eqref{eq:student_teacher_approx} and discuss the assumptions in more detail. \\

\noindent With ConfiG we aim to generate augmentations with high confidence of the teacher and low confidence of the student. This corresponds to $\mathcal{Q}$ having the following support.
\begin{equation}
    \text{supp}(\mathcal{Q})=\{\textbf{x}| \; \textbf{t}_{y_{\textbf{t}}}(\textbf{x})\ge\tau \; \text{and} \; \textbf{f}_{y_{\textbf{t}}}(\textbf{x})\le\sigma<\tau \}
\end{equation}
where $y_{\textbf{t}}(\textbf{x}) = \underset{k\in[L]}{\text{argmax }} \textbf{t}_k(\textbf{x})$ is the teacher’s predicted class. Assume that the data-generating distribution $\textbf{p}^*$ is concentrated around the teacher predictions on $\mathcal{Q}$ such that there exists $\eta\in[0,1)$ with
\begin{equation}
    \textbf{p}^*_{y_\textbf{t}}(\textbf{x})\geq 1 - \eta\;\;\text{for all }\textbf{x}\in\text{supp}(\mathcal{Q})
\end{equation}
If $\underset{y\ne y_{\textbf{t}}}{\text{sup}}\bigl| \textbf{log}\frac{\textbf{t}_y(\textbf{x})}{\textbf{f}_y(\textbf{x})}\bigr|\leq C$ for all $\textbf{x}\in\text{supp}(\mathcal{Q})$, we can bound the difference in risk between the teacher and student on $\mathcal{Q}$
\begin{align}
\mathcal{R}_\mathcal{Q}(\textbf{f})-\mathcal{R}_\mathcal{Q}(\textbf{t})
& = \mathbb{E}_{x\sim\mathcal{Q}_\textbf{x}}\!\Biggl[\sum_{y} p^*_y(\textbf{x})\log\!\frac{t_y(\textbf{x})}{f_y(\textbf{x})}\Biggr] \\
& =  \mathbb{E}_{x\sim\mathcal{Q}_\textbf{x}}\!\Biggl[p^*_{y_{\textbf{t}}}(\textbf{x})\log\frac{t_{y_{\textbf{t}}}(\textbf{x})}{f_{y_{\textbf{t}}}(\textbf{x})}+\sum_{y\ne y_{\textbf{t}}} p^*_y(\textbf{x})\log\!\frac{t_y(\textbf{x})}{f_y(\textbf{x})}\Biggr] \\
& \ge \mathbb{E}_{x\sim\mathcal{Q}_x}\!\Bigl[p^*_{y_{\textbf{t}}(\textbf{x})}(\textbf{x})\log\frac{\tau}{\sigma} - (1-p^*_{y_{\textbf{t}}(\textbf{x})}(\textbf{x}))\,C\Bigr] \\
& \ge (1-\eta)\log\!\frac{\tau}{\sigma} - \eta C.
\end{align}
Here $\mathcal{Q}_\textbf{x}$ denotes the marginal distribution of $\textbf{x}$ on $\mathcal{Q}$. With $\log\frac{\tau}{\sigma}$ sufficiently large, this satisfies assumption~\eqref{eq:student_teacher_approx}. \\
Next we discuss the individual assumptions of Proposition~\ref{prop:1}. $\Omega \;=\;\mathcal{R}_{test}(\textbf{f}) - \mathcal{R}_{train}(\textbf{f})>0$ is a realistic scenario under covariate shift as learning spurious features instead of causal features typically results in low training but high test risk, in particular for the auxiliary students used to generate ConfiG augmentations.
We assume
\begin{align}
KL\!\bigl(\textbf{p}^*(\textbf{x})\,\|\,\textbf{t}(\textbf{x})\bigr) \leq \kappa_t\,\text{on the support of } \mathcal{P}_{\text{test}},\mathcal{P}_{\text{train}},\mathcal{Q}\\
0<\mathcal{R}_{\mathrm{train}}(\textbf{f})-\mathcal{R}_{\mathrm{train}}(\textbf{t}) \le \epsilon_{train}\,\text{on the support of } \mathcal{P}_{\text{train}}
\\
\mathcal{R}_\mathcal{Q} (\textbf{f}) -\mathcal{R}_\mathcal{Q}(\textbf{t})\geq \epsilon_{aug}>0\,\text{on the support of } \mathcal{Q}.
\end{align}
The first bound reflects that the teacher approximates the data-generating distribution $\textbf{p}^*$. This requires the teacher to be both accurate and well-calibrated. These two properties have been identified as favorable characteristics for a teacher model in knowledge distillation by prior work~\cite{statistical,cutmixpick}. The second bound requires the student to achieve similar training loss as the teacher model.
This is the goal when distilling the student on the original training distribution.
$\Delta_{\textbf{p}}$ measures how much the risk of the true data-generating distribution changes under the augmentations.
The third condition ensures that the data augmentations are more challenging for the student than for the teacher.
As described above, under realistic assumptions this can be achieved with ConfiG through large $\tau$ and small $\sigma$.
The condition on $\Theta$ ensures that the confidence gap is sufficiently large such that the augmented samples bring $\mathcal{P}_{\text{train}}$ closer to $\mathcal{P}_{\text{test}}$.
The final condition on $\alpha$ reflects that sampling too many augmented images can bias $\mathcal{P}_{aug}$ towards $\mathcal{Q}$, which in turn harms $\Omega$.
Both conditions on $\Theta$ and $\alpha$ can be controlled through the type and number of augmented images in practice such that the proposition does not apply to an empty set.\\

\noindent Overall, we highlight that the individual assumptions are approximately satisfied in a common setting with $\ell$ being the cross-entropy loss, a strong teacher which is accurate both on $\mathcal{P}_{\text{train}}$ and $\mathcal{P}_{\text{test}}$, a well-trained student and augmentations generated by ConfiG through optimizing $\mathcal{L}_{ConfiG}$. However, there are two main limitations for practical use of our result: first, it would be very difficult to characterize the distribution $\mathcal{Q}$ purely from a discrete set of ConfiG augmentations, and second, even if this was feasible, estimating the constants in order to get $\alpha$ is infeasible. In addition, ConfiG augmentations are added to the discrete training set which can influence both $\Omega$ and $\Psi$ that jointly constitute the generalization error of the student.

\section{Experimental Details}\label{app:hyperparameter}

In this section we summarize the hyperparameters used for training the student models, generating ConfiG augmentations, the non-synthetic data augmentations and the details for the teacher inference. Code is available at \url{https://github.com/boschresearch/ConfidenceGuidedAugmentation}.

\subsubsection{Training Hyperparameters} All pre-trained model weights were downloaded from the timm library~\cite{rw2019timm}.
The student models are trained for 300 epochs using the AdamW optimizer~\cite{adamw} with a constant learning rate of $10^{-5}$. 
For all experiments on the group shift datasets we use a batchsize of 128. On ImageNet-100 we use 2 GPUs with a local batchsize of 128 resulting in a global batchsize of 256. The final checkpoint is evaluated to obtain the test results.
For the group shift datasets we generate synthetic augmentations once with every method and repeat the student trainings six times. We report the mean value and the \textcolor{gray}{standard deviation (following $\pm$)} in the tables.
For ImageNet-100, student trainings are performed once per method due to the larger computational cost. We follow VanillaKD~\cite{VanillaKD} and use a weight decay of 0.02, for the group shift datasets we use no weight decay. 
Tab.~\ref{tab:hyperparams_training} provides an overview over the training hyperparameters.

\begin{table*}[tb]
\centering
\begin{threeparttable}
\caption{Training hyperparameters. }
\label{tab:hyperparams_training}
\begin{tabular}{@{}lcc@{}}
\toprule
\textbf{Hyperparameter \hspace{.6cm}} & \textbf{CelebA-HQ, SpuCo Birds, BAR\hspace{.6cm}} & \textbf{ImageNet-100} \\
\midrule
Epochs & 300 & 300 \\
Learning Rate & $1 \times 10^{-5}$ & $1 \times 10^{-5}$ \\
Weight Decay & 0 & 0.02 \\
Batchsize & 128 & 256 \\
LR Schedule & Constant & Constant \\
Optimizer & AdamW & AdamW \\
\bottomrule
\end{tabular}
\end{threeparttable}
\end{table*}

\subsubsection{Training and Test Data Selection} 
\begin{figure}[tbp]
    \centering
    \hspace{-.5cm}
    \includegraphics[width=.8\columnwidth]{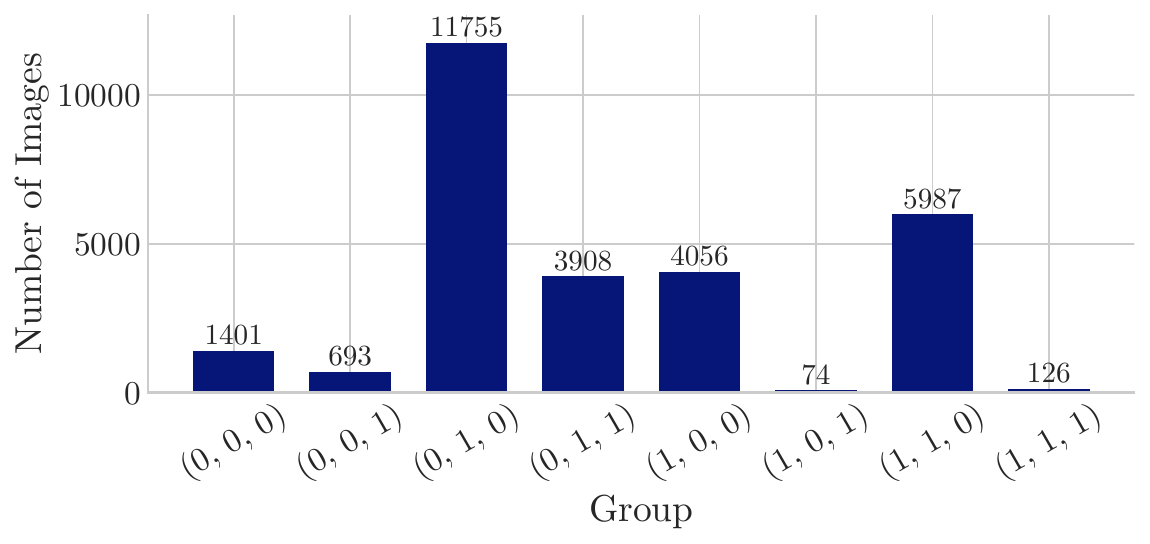}
    %\vskip-0.2cm
    \caption{\textbf{Group Distribution in the Custom Test Set of CelebA-HQ.} The first label refers to gender (0: female, 1: male), the second label to age (0: non-young, 1: young) and the third label to hair color (0: non-blond, 1: blond).
    }
    \label{fig:celeba_test_dist}
    %\vskip-0.2cm
    %\vspace{-2mm}
\end{figure}

For SpuCo Birds and BAR, we randomly subsample images from the standard train set and test on the standard test set. On CelebA-HQ the test set is substantially smaller than the train set and some groups contain very few examples. Thus, we select our training data from the original CelebA-HQ test set and evaluate on the original train dataset. All our student models were pre-trained on ImageNet-1k or 21k and thus have not encountered any images from CelebA-HQ during pre-training. Fig.~\ref{fig:celeba_test_dist} shows the distribution of images across the eight subgroups in our test set.\\
ImageNet-100 contains 100 images for all classes in the Spurious ImageNet dataset~\cite{spurious_imagenet}. We select the top-100 images per class from the original ImageNet-1k~\cite{imagenet} training set with the highest cosine similarity with respect to the spurious components used to construct Spurious ImageNet.

\subsubsection{ConfiG Hyperparameters} 
\begin{figure}[bt]
    %\vskip-.35cm
    \centering
    \hspace{-.5cm}
    \includegraphics[width=.8\columnwidth]{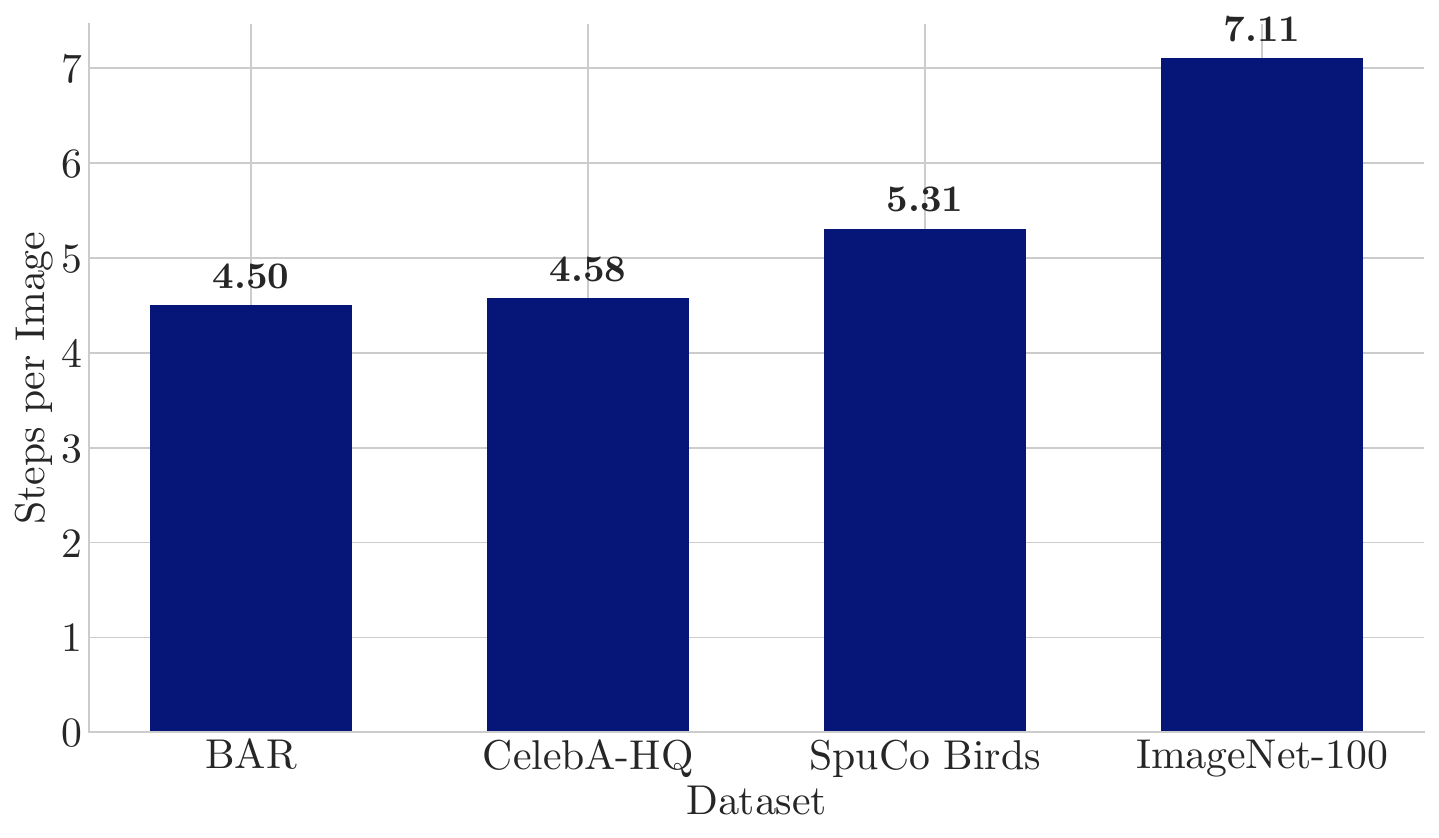}
    %\vskip-0.2cm
    \caption{Average Number of steps for ConfiG augmentations per dataset. The first step is used for inter-class mixing, the consecutive steps for latent optimization.
    }
    \label{fig:steps_per_im}
    %\vskip-0.2cm
    %\vspace{-2mm}
\end{figure}

The latent optimization of ConfiG is performed using the Adam~\cite{adam} optimizer and a learning rate of 0.01 for the conditioning and null-text and a learning rate of 0.001 for the initial latent. With Stable Diffusion 1.4 we perform at most 10 optimization steps or stop when the loss from Eq.~\eqref{eq:loss} reaches a value below 0.05. If this stopping criterion is not reached we use the image with the lowest loss from the optimization trajectory. The image generation is performed with guidance scale 3.0 and 20 inference steps following DiG-IN~\cite{dig_in}. 
The resulting images have resolution 512$\times$512.
The auxiliary student used for optimizing the ConfiG loss is trained with standard augmentations for the group shift datasets and A1 augmentations for ImageNet-100.
In Fig.~\ref{fig:steps_per_im}, we show the average number of steps for each of the four datasets. The number of steps includes the first step for inter-class mixing followed by the optimization steps. As prompt template we use ``a realistic photo of $\{$classname$\}$''. For CelebA-HQ we use the classnames ``female face'' and ``male face'' and for SpuCo Birds we use ``landbird'' and ``waterbird''. For ImageNet-100 we use the short classnames used in the work of Spurious ImageNet~\cite{spurious_imagenet}. For BAR, we nominalize the classnames to ``climber'', ``diver'', ``fisher'', ``pole vaulter'', ``motorsports race'' and ``thrower'' to fit this prompt template. For prompt-to-prompt editing with Stable Diffusion 1.4 we add the suffix ``in front of a background'' to the prompt to enable the editing to distinguish between the class object and the background. We use the implementation provided by DiG-IN~\cite{dig_in} which is a modified version of the original prompt-to-prompt editing. Null-text inversion is performed with the guidance scale 7.5. For the group shift datasets, we invert with 20 DDIM steps. For ImageNet-100, we invert with 25 steps, of which the first 20 steps are used as initialization for the null-tokens. We found this to reduce the number of ConfiG optimization steps and improve downstream performance on ImageNet-100 where the semantic shifts between different classes during inter-class mixing are larger than for the domain-specific group shift datasets. The image transformation for null-text inversion is taken from DiG-IN and consists of a resize and center-crop that crops to 80$\%$ of the original image along the shorter side. The original images in all Figures of this paper are shown after this transformation. To increase diversity when generating multiple augmentations for the same image, we add a small random perturbation to the initial latent on the group shift datasets. Following DiG-IN~\cite{dig_in}, with Stable Diffusion 1.4 we add a regularization term for the background which uses HQ-SAM~\cite{sam_hq} to distinguish between background and foreground. For the Hyper-SD single-step model~\cite{hyper-sd} used in App.~\ref{sec:hyper_sd} we perform up to 25 optimization steps with a learning rate of 0.1 due to the faster encoding time and keep the same early stopping criterion. In addition, we repeat the optimization process up to three times if the loss does not reach a value below 0.05 beforehand. For Hyper-SD we leverage IP-Adapter guidance instead of null-text inversion and prompt-to-prompt editing. The IP-Adapter scale is set to 0.9 for SpuCo Birds and 0.8 for BAR and CelebA. Augmentations with Hyper-SD are generated at resolution 1024$\times$1024.

\subsubsection{Non-Synthetic Augmentations} In Tab.~\ref{tab:hyperparams_augmentations} we list the hyperparameters used for the non-synthetic data augmentations. For CutMix and MixUp we use the default hyperparameters from torchvision~\cite{torchvision}. The A1 augmentations for ImageNet-100 are the same as in VanillaKD~\cite{VanillaKD,a1_aug}. Note that for our experiments, we fully replace a real image with a synthetic data augmentation with probability $\alpha$ before applying non-synthetic augmentations. Following Diff-Mix, the synthetic samples are drawn from a global random pool~\cite{survery2026,Diff-Mix}. We use this setup for all baselines for a consistent comparison. In alternate work such as DiffuseMix~\cite{diffuse}, real images are only partially replaced with synthetic augmentations.

\begin{table*}[bt]
\centering
\begin{threeparttable}
\caption{\textbf{Hyperparameter Configurations for the Non-Synthetic Data Augmentation Strategies.} used in our experiments on group shift datasets. Checkmarks (\cmark) and crosses (\xmark) indicate the application of a specific augmentation. RRC abbreviates \underline{R}andom \underline{R}esized \underline{C}rops.}
\label{tab:hyperparams_augmentations}
\renewcommand{\arraystretch}{1.2} % Adds a bit more vertical spacing for readability
\begin{tabular}{@{}lccccc@{}}
\toprule
 & \thead{Default} & \thead{Stronger RRC \\ + H. Flips} & \thead{+ AutoAugment} & \thead{+ MixUp} & \thead{+ CutMix} \\
\midrule
RRC Scale & (0.9, 1.0) & (0.08, 1.0) & (0.08, 1.0) & (0.08, 1.0) & (0.08, 1.0) \\
RRC Ratio & (3/4, 4/3) & (3/4, 4/3)  & (3/4, 4/3)  & (3/4, 4/3)  & (3/4, 4/3)  \\
H. Flip Probability & 0.0 & 0.5 & 0.5 & 0.5 & 0.5 \\
\midrule
AutoAugment & \xmark & \xmark & \scriptsize{rand-m7-mstd0.5-inc1} & \xmark & \xmark \\
MixUp & \xmark & \xmark & \xmark & \cmark & \xmark \\
CutMix & \xmark & \xmark & \xmark & \xmark & \cmark \\
\bottomrule
\end{tabular}
\end{threeparttable}
%\vspace{-.4cm}
\end{table*}

\subsubsection{Teacher Model Inference} On CelebA-HQ and BAR we obtain the teacher logits using the cosine similarity between class-specific text embeddings and the image embeddings. The prompt used for the class-specific text embeddings is ``a photo of $\{$classname$\}$''. The logits are multiplied by a CLIP-specific factor of 100 for temperature scaling before applying the softmax to obtain the teacher output distribution. Since the classnames for SpuCo Birds (waterbird/landbird) contain the spurious feature (water/land), directly using these names for zero-shot inference of a CLIP model yields poor performance. Thus, we use the individual names of the bird species which were in turn used to construct the SpuCo Birds dataset to obtain fine-grained text-to-image similarities. The maximum logit from all waterbird species and the maximum logit from all landbird species are subsequently taken as the logits for the two classes in SpuCo Birds. On ImageNet-100 we use a BeiTv2-B~\cite{beitv2} teacher trained on ImageNet-1k\footnote{\url{https://huggingface.co/timm/beitv2_base_patch16_224.in1k_ft_in1k}} and mask the output of the teacher to the 100 classes in our training dataset.

\subsubsection{Model-Centric Training} For pure ERM, LfF and uLA we use the same pre-trained student model checkpoint and learning rate as for knowledge distillation in Sec.~\ref{sec:group_shift_datasets} with one ConfiG augmentation per image. 
With uLA we use the hyperparameters $\tau=0.5$ and $\eta=1.5$ for CelebA-HQ which were used for CelebA in the original work. For SpuCo Birds we leverage $\tau=0.1$ and $\eta=3.0$ which the original work applied in case of the Waterbirds~\cite{distr_robust} dataset. For BAR we use $\tau=1.0$ and $\eta=1.0$ which were used for the cCIFAR10~\cite{cifar,corrupted} dataset.
For the generalized cross-entropy loss in LfF we use $q=0.7$. 
Since LfF uses an index-based EMA updates for the relative difficulty scores we concatenate the real data and ConfiG augmentations into one joint training set and reduce the overall number of epochs to 150.
For a consistent comparison, we evaluate the final model checkpoints to obtain the model performance for all model-centric training. 
Since we assume the covariate shift is unknown and the generated augmentations do not provide additional labels for assigning samples to groups, we omit methods such as DRO~\cite{distr_robust} that require predefined group information.

\section{Computational Cost of ConfiG and Baseline Multipliers}\label{app:cost}

Tab.~\ref{tab:augmentation_cost} lists the computational cost of generating a single synthetic augmentation for each method. To enable comparisons at approximately equal generation cost, we introduce multipliers that determine how many augmentations are generated relative to ConfiG. One-time preprocessing costs, such as diffusion-model fine-tuning, suffix generation or null-text inversion are excluded.
For baselines we use the default hyperparameters in the respective codebases unless specified differently in App.~\ref{app:baseline summary}.
For ConfiG, we estimate the average FLOP cost per augmentation by combining the cost of the initial inter-class mixing step with the cost of performing one optimization and denoising step multiplied by the average number of optimization steps. Fig.~\ref{fig:steps_per_im} reports the total number of steps for the experiments with the ViT-T student in Sec.~\ref{sec:experiments}. The total number of steps includes inter-class mixing (first step) as well as the subsequent optimization steps. 
All measurements were obtained using the PyTorch FLOP counter with batchsize 1. Due to memory limitations, we could only measure the FLOPs of the forward passes for ConfiG directly. The FLOP cost of the backward passes is therefore estimated as the cost of the corresponding forward passes multiplied by factor 2. The reported cost for ConfiG depends on this factor. A potentially larger factor would increase the budget allocated to all baselines and our approximations are therefore conservative against the baselines.
For simplicity, we use one approximate factor for all group shift datasets and one for ImageNet-100.

\begin{table}[tbp]
\centering
\scriptsize
\begin{threeparttable}
\caption{\textbf{Approximate Computational Cost per Augmentation Across Different Augmentation Methods.} We report the estimated FLOPs required to generate a single augmentation (measured on the BAR dataset) together with the effective augmentation multiplier used for a comparison under approximately equal FLOPs. \vspace{-.2cm}}
\label{tab:augmentation_cost}
\begin{tabular}{@{}lcc@{}}
\toprule
\textbf{Method} & \textbf{Approximate Cost per Aug. Generation} & \textbf{Multiplier} \\
\midrule
ConfiG & $2.8\times10^{14}$ (first step for inter-class mixing) & $1\times$ \\
 & $+$ $1.2\times10^{14}$ (every additional step) &  \\
ActGen & $1.0\times10^{14}$ & $8\times$ (CelebA-HQ, SpuCo Birds, BAR),\\
 &  & $10\times$ (ImageNet-100) \\
Diff-Mix & $2.7\times10^{13}$ & $28\times$ (CelebA-HQ, SpuCo Birds, BAR), \\
 &  & $36 \times$ (ImageNet-100) \\
Diff-II & $4.3\times10^{13}$ & $16\times$ (CelebA-HQ, SpuCo Birds, BAR),\\
 &  & $22\times$ (ImageNet-100)\\
DistDiff & $6.0\times10^{13}$ & $12\times$ (CelebA-HQ, SpuCo Birds, BAR), \\
 &  & $16\times$ (ImageNet-100) \\
DA-Fusion & $5.7\times10^{13}$ & $12\times$ (CelebA-HQ, SpuCo Birds, BAR), \\
 & & $16\times$ (ImageNet-100)\\
GIF & $4.9\times10^{13}$ &  $16\times$ (CelebA-HQ, SpuCo Birds, BAR), \\
 & $8.1\times10^{13}$ & $12 \times$ (ImageNet-100) \\
\bottomrule
\end{tabular}
\end{threeparttable}
\end{table}

\begin{table*}[tb]
\centering
\begin{threeparttable}
\caption{Diffusion-based data augmentations combined with stronger random resized crops and horizontal flips in the setting of Tab.~\ref{tab:group_shift}.}
\label{tab:rrc_full results}
\begin{tabular}{@{}lcccccc@{}}
\toprule
& \multicolumn{3}{c}{\textbf{CelebA-HQ}} & \multicolumn{2}{c}{\textbf{SpuCo Birds}} & \textbf{BAR} \\
\cmidrule(lr){2-4} \cmidrule(lr){5-6} \cmidrule(lr){7-7}
\textbf{Method} & \textbf{SMA} & \textbf{GMA} & \textbf{WGA} & \textbf{SMA} & \textbf{WGA} & \textbf{SMA} \\
\midrule
Real Data & 97.17\,\stdv{0.66} & 91.81\,\stdv{0.77} & 69.97\,\stdv{7.19} & 60.58\,\stdv{0.79} & 17.63\,\stdv{2.77} & 52.17\,\stdv{3.98} \\
\midrule
GIF & 97.89\,\stdv{0.13} & 92.54\,\stdv{0.60} & 72.56\,\stdv{4.27} & 66.84\,\stdv{1.11} & 29.46\,\stdv{2.36} & 57.70\,\stdv{2.80} \\
DistDiff & 97.56\,\stdv{0.19} & 93.53\,\stdv{0.89} & 79.63\,\stdv{5.37} & 65.88\,\stdv{1.67} & 23.23\,\stdv{2.77} & 57.98\,\stdv{2.99} \\
DA-Fusion & 97.69\,\stdv{0.32} & \underline{94.32}\,\stdv{0.38} & \underline{83.67}\,\stdv{2.20} & \underline{72.10}\,\stdv{1.68} & \underline{33.87}\,\stdv{6.55} & 63.86\,\stdv{3.94} \\
Diff-Mix & 97.69\,\stdv{0.38} & 93.58\,\stdv{1.02} & 72.62\,\stdv{6.69} & 70.28\,\stdv{3.28} & 33.40\,\stdv{6.39} & 64.91\,\stdv{1.16} \\
Diff-II & 97.57\,\stdv{0.12} & 93.23\,\stdv{0.51} & 75.13\,\stdv{3.81} & 64.06\,\stdv{1.20} & 25.50\,\stdv{2.21} & \underline{64.98}\,\stdv{1.89} \\
ActGen & \underline{97.92}\,\stdv{0.15} & 93.24\,\stdv{0.91} & 74.47\,\stdv{4.90} & 64.55\,\stdv{2.72} & 25.10\,\stdv{4.01} & 59.86\,\stdv{1.81} \\
ConfiG & \textbf{98.42}\,\stdv{0.05} & \textbf{96.72}\,\stdv{0.30} & \textbf{90.98}\,\stdv{1.40} & \textbf{73.78}\,\stdv{1.60} & \textbf{42.10}\,\stdv{5.00} & \textbf{65.39}\,\stdv{1.91} \\
\bottomrule
\end{tabular}
\end{threeparttable}
\end{table*}
\begin{table*}[tb]
\centering
\begin{threeparttable}
\caption{Diffusion-based data augmentations combined with stronger random resized crops, horizontal flips and AutoAugment in the setting of Tab.~\ref{tab:group_shift}.}
\label{tab:aa_full results}
\begin{tabular}{@{}lcccccc@{}}
\toprule
& \multicolumn{3}{c}{\textbf{CelebA-HQ}} & \multicolumn{2}{c}{\textbf{SpuCo Birds}} & \textbf{BAR} \\
\cmidrule(lr){2-4} \cmidrule(lr){5-6} \cmidrule(lr){7-7}
\textbf{Method} & \textbf{SMA} & \textbf{GMA} & \textbf{WGA} & \textbf{SMA} & \textbf{WGA} & \textbf{SMA} \\
\midrule
Real Data & 97.82\,\stdv{0.11} & 92.50\,\stdv{0.67} & 72.22\,\stdv{3.76} & 65.46\,\stdv{0.80} & 25.17\,\stdv{1.99} & 61.65\,\stdv{2.99} \\
\midrule
GIF & 97.81\,\stdv{0.20} & 91.97\,\stdv{0.73} & 69.97\,\stdv{2.72} & 67.97\,\stdv{2.11} & 27.73\,\stdv{5.29} & 62.28\,\stdv{3.24} \\
DistDiff & 97.91\,\stdv{0.24} & 93.54\,\stdv{0.58} & 79.03\,\stdv{3.99} & 70.73\,\stdv{2.11} & 29.57\,\stdv{3.03} & 62.82\,\stdv{1.18} \\
DA-Fusion & 97.92\,\stdv{0.27} & \underline{94.33}\,\stdv{0.37} & \underline{82.78}\,\stdv{2.86} & \underline{73.74}\,\stdv{1.41} & 33.50\,\stdv{3.00} & 65.88\,\stdv{1.46} \\
Diff-Mix & 97.86\,\stdv{0.21} & 93.60\,\stdv{0.81} & 74.21\,\stdv{5.61} & 72.63\,\stdv{2.24} & \underline{36.50}\,\stdv{7.86} & 67.89\,\stdv{2.27} \\
Diff-II & 97.72\,\stdv{0.08} & 93.18\,\stdv{0.61} & 76.06\,\stdv{5.20} & 68.15\,\stdv{2.02} & 30.37\,\stdv{5.68} & \underline{68.07}\,\stdv{0.40} \\
ActGen & \underline{97.95}\,\stdv{0.33} & 93.99\,\stdv{0.65} & 80.63\,\stdv{3.82} & 71.09\,\stdv{1.18} & 33.20\,\stdv{2.12} & 64.48\,\stdv{0.82} \\
ConfiG & \textbf{98.53}\,\stdv{0.05} & \textbf{96.55}\,\stdv{0.36} & \textbf{89.45}\,\stdv{1.59} & \textbf{78.62}\,\stdv{1.61} & \textbf{50.53}\,\stdv{5.34} & \textbf{69.78}\,\stdv{0.70} \\
\bottomrule
\end{tabular}
\end{threeparttable}
\end{table*}
\begin{table*}[tb]
\centering
\begin{threeparttable}
\caption{Diffusion-based data augmentations combined with stronger random resized crops, horizontal flips and MixUp in the setting of Tab.~\ref{tab:group_shift}.}
\label{tab:mixup_full results}
\begin{tabular}{@{}lcccccc@{}}
\toprule
& \multicolumn{3}{c}{\textbf{CelebA-HQ}} & \multicolumn{2}{c}{\textbf{SpuCo Birds}} & \textbf{BAR} \\
\cmidrule(lr){2-4} \cmidrule(lr){5-6} \cmidrule(lr){7-7}
\textbf{Method} & \textbf{SMA} & \textbf{GMA} & \textbf{WGA} & \textbf{SMA} & \textbf{WGA} & \textbf{SMA} \\
\midrule
Real Data & 97.66\,\stdv{0.43} & 93.15\,\stdv{1.08} & 74.35\,\stdv{8.53} & 68.99\,\stdv{0.71} & 37.20\,\stdv{1.31} & 61.70\,\stdv{1.71} \\
\midrule
GIF & 97.76\,\stdv{0.08} & 92.66\,\stdv{0.58} & 73.94\,\stdv{4.33} & 71.96\,\stdv{0.86} & 37.37\,\stdv{4.53} & 60.04\,\stdv{2.65} \\
DistDiff & 97.84\,\stdv{0.15} & 92.96\,\stdv{0.79} & 74.74\,\stdv{5.15} & 72.61\,\stdv{2.35} & 39.40\,\stdv{5.96} & 63.23\,\stdv{2.56} \\
DA-Fusion & 97.85\,\stdv{0.12} & 94.27\,\stdv{0.78} & 82.18\,\stdv{4.24} & \underline{74.87}\,\stdv{2.16} & 41.23\,\stdv{7.16} & 64.86\,\stdv{2.10} \\
Diff-Mix & \underline{98.06}\,\stdv{0.08} & 94.38\,\stdv{0.60} & 77.65\,\stdv{3.60} & 73.19\,\stdv{1.28} & \underline{43.00}\,\stdv{4.93} & 65.62\,\stdv{1.10} \\
Diff-II & 97.69\,\stdv{0.15} & 93.02\,\stdv{0.68} & 72.49\,\stdv{5.42} & 69.03\,\stdv{2.29} & 35.47\,\stdv{5.02} & \underline{66.56}\,\stdv{1.78} \\
ActGen & 97.72\,\stdv{0.23} & \underline{94.48}\,\stdv{0.45} & \underline{83.78}\,\stdv{2.69} & 72.81\,\stdv{2.55} & 39.03\,\stdv{6.10} & 61.95\,\stdv{1.62} \\
ConfiG & \textbf{98.42}\,\stdv{0.11} & \textbf{96.79}\,\stdv{0.48} & \textbf{89.95}\,\stdv{1.38} & \textbf{77.99}\,\stdv{1.97} & \textbf{51.57}\,\stdv{5.73} & \textbf{68.81}\,\stdv{1.86} \\
\bottomrule
\end{tabular}
\end{threeparttable}
\end{table*}
\begin{table*}[t]
\centering
\begin{threeparttable}
\caption{Diffusion-based data augmentations combined with stronger random resized crops, horizontal flips and CutMix in the setting of Tab.~\ref{tab:group_shift}.}
\label{tab:cutmix_full results}
\begin{tabular}{@{}lcccccc@{}}
\toprule
& \multicolumn{3}{c}{\textbf{CelebA-HQ}} & \multicolumn{2}{c}{\textbf{SpuCo Birds}} & \textbf{BAR} \\
\cmidrule(lr){2-4} \cmidrule(lr){5-6} \cmidrule(lr){7-7}
\textbf{Method} & \textbf{SMA} & \textbf{GMA} & \textbf{WGA} & \textbf{SMA} & \textbf{WGA} & \textbf{SMA} \\
\midrule
Real Data & 97.74\,\stdv{0.26} & 93.17\,\stdv{0.77} & 70.24\,\stdv{4.59} & 70.05\,\stdv{2.15} & 36.63\,\stdv{4.29} & 58.21\,\stdv{2.61} \\
\midrule
GIF & 97.94\,\stdv{0.14} & 93.31\,\stdv{1.00} & 73.41\,\stdv{6.46} & 68.68\,\stdv{0.94} & 32.20\,\stdv{4.23} & 55.96\,\stdv{2.84} \\
DistDiff & 98.06\,\stdv{0.09} & 93.94\,\stdv{0.46} & 76.19\,\stdv{2.75} & 69.70\,\stdv{1.84} & 32.93\,\stdv{6.66} & 61.47\,\stdv{1.78} \\
DA-Fusion & 98.11\,\stdv{0.09} & \underline{94.43}\,\stdv{0.62} & \underline{79.37}\,\stdv{3.55} & \underline{74.18}\,\stdv{1.12} & 37.97\,\stdv{3.35} & \underline{63.12}\,\stdv{2.03} \\
Diff-Mix & 97.77\,\stdv{0.07} & 93.50\,\stdv{0.40} & 71.43\,\stdv{2.41} & 71.48\,\stdv{2.37} & \underline{38.97}\,\stdv{3.55} & 62.31\,\stdv{1.69} \\
Diff-II & 97.75\,\stdv{0.26} & 93.08\,\stdv{1.11} & 70.50\,\stdv{6.20} & 69.13\,\stdv{1.76} & 34.37\,\stdv{2.75} & 62.08\,\stdv{1.72} \\
ActGen & \underline{98.12}\,\stdv{0.10} & 94.06\,\stdv{1.20} & 76.98\,\stdv{7.13} & 70.59\,\stdv{2.52} & 34.60\,\stdv{6.50} & 59.00\,\stdv{1.22} \\
ConfiG & \textbf{98.33}\,\stdv{0.21} & \textbf{96.14}\,\stdv{0.67} & \textbf{87.90}\,\stdv{2.80} & \textbf{77.66}\,\stdv{0.52} & \textbf{53.10}\,\stdv{3.42} & \textbf{65.34}\,\stdv{1.75} \\
\bottomrule
\end{tabular}
\end{threeparttable}
\end{table*}
\section{Full Results with Stronger Standard Augmentations}\label{app:stronger_aug_full}
Results for combining the diffusion-based data augmentations with stronger non-synthetic data augmentations are presented in Tab.~\ref{tab:rrc_full results},~\ref{tab:aa_full results},~\ref{tab:mixup_full results} and~\ref{tab:cutmix_full results}. We find that the performance on the BAR dataset with only real data and MixUp or AutoAugment surpasses the results with ConfiG and only random resized crops from Tab.~\ref{tab:group_shift}. On CelebA-HQ and SpuCo Birds, ConfiG with only random resized crops outperforms all non-synthetic data augmentation in GMA and WGA. The best overall performances can be achieved by combining ConfiG with stronger random resized crops, AutoAugment or MixUp on CelebA-HQ and ConfiG with AutoAugment on BAR. On SpuCo Birds, the highest student accuracies are achieved by combining ConfiG with either MixUp or CutMix.

\begin{table*}[t]
\centering
\begin{threeparttable}
\caption{\textbf{ConfiG with a Single-Step Diffusion Model.} We use SDXL Hyper-SD~\cite{hyper-sd} to demonstrate that ConfiG can be used with single-step diffusion models to improve the student performance under covariate shift.}
\label{tab:hyper_sd}
\begin{tabular}{@{}lcccccc@{}}
\toprule
& \multicolumn{3}{c}{\textbf{CelebA-HQ}} & \multicolumn{2}{c}{\textbf{SpuCo Birds}} & \textbf{BAR} \\
\cmidrule(lr){2-4} \cmidrule(lr){5-6} \cmidrule(lr){7-7}
\textbf{Method} & \textbf{SMA} & \textbf{GMA} & \textbf{WGA} & \textbf{SMA} & \textbf{WGA} & \textbf{SMA} \\
\midrule
Real Data & 93.98\,\stdv{1.68} & 86.89\,\stdv{2.48} & 53.44\,\stdv{8.12} & 56.53\,\stdv{2.40} & 12.97\,\stdv{6.65} & 37.69\,\stdv{3.58} \\
\midrule
SD 1.4 & \textbf{97.76}\,\stdv{0.25} & \textbf{95.67}\,\stdv{0.55} & \textbf{88.15}\,\stdv{2.10} & \textbf{73.38}\,\stdv{1.28} & \textbf{39.50}\,\stdv{4.53} & \textbf{58.66}\,\stdv{3.76} \\

Hyper-SD & 97.55\,\stdv{0.28} & 94.17\,\stdv{0.87} & 81.51\,\stdv{5.96} & 64.75\,\stdv{3.26} & 31.43\,\stdv{7.23} & 53.08\,\stdv{4.57} \\
\bottomrule
\end{tabular}
\end{threeparttable}
%\vspace{-.3cm}
\end{table*}
\section{Single-Step Diffusion Model for Synthesis}\label{sec:hyper_sd}
Distilled diffusion models such as single-step or few-step consistency models~\cite{hyper-sd,lcm} offer faster image generation in comparison to Stable Diffusion 1.4 or Stable Diffusion XL models. Distilled single-step consistency models typically employ specialized schedulers such as TCD~\cite{tcd} instead of the DDPM~\cite{ddpm} or DPMsolver++~\cite{DPMsolver++} scheduler which the strongest baselines Diff-Mix, Diff-II and ActGen utilize. In Tab.~\ref{tab:hyper_sd} we demonstrate that ConfiG can be carried out with a single-step Hyper-SD~\cite{hyper-sd} model distilled from Stable Diffusion XL. Since Hyper-SD is not compatible with null-text inversion and prompt-to-prompt editing, we leverage IP-Adapter~\cite{ipadapter} with the original image and the target class prompt as guidance for inter-class mixing.
On all three group shift datasets, training with ConfiG augmentations from this setup consistently improves the student performance over training on real data only. On CelebA-HQ, Hyper-SD achieves performances close to those of Stable Diffusion 1.4 with null-text inversion and prompt-to-prompt editing with a decrease of 0.2$\%$ in SMA and 1.5$\%$ in GMA.
On SpuCo Birds and BAR larger differences in SMA emerge.
A potential reason for this observation could be the larger domain gap between the real images and the synthetic images generated by Hyper-SD with IP-Adapter guidance~\cite{sim2real_consistency}.  This is reflected in our theoretical analysis in App.~\ref{sec:theoretical}. Similarly, Li et al.~\cite{survery2026} observe that using latent consistency models~\cite{lcm} instead of Stable Diffusion models results in a slight drop in accuracy.

\section{More Augmentations per Image}\label{app:more_auf}
\begin{figure}[tp!]
    \centering
    % \hspace{-.5cm}
    \includegraphics[width=.99\columnwidth]{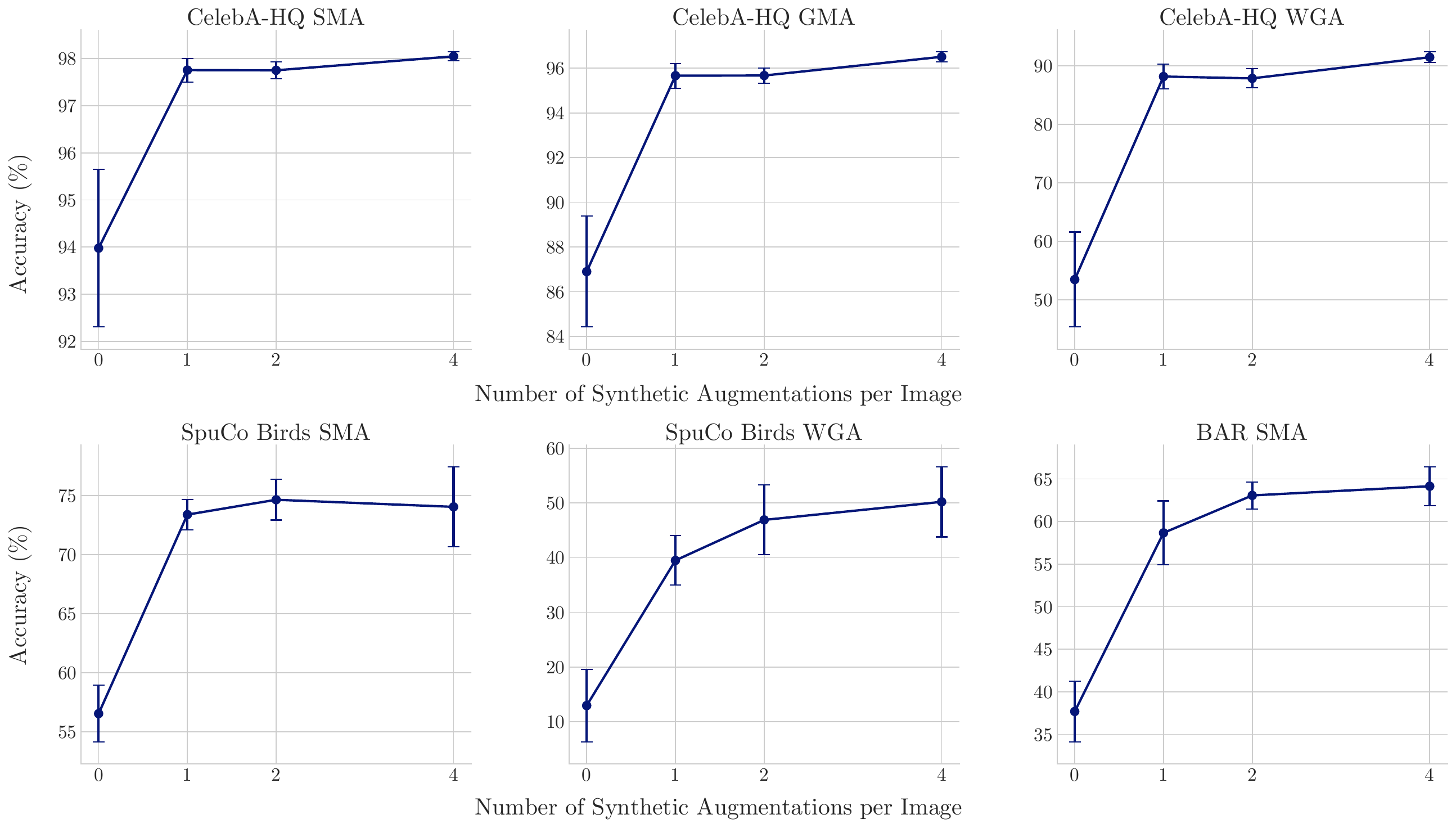}
    %\vskip-0.2cm
    \caption{\textbf{Scaling the Number of ConfiG Augmentations per Image.} Using more than one ConfiG augmentation per image further improves the student performance. Four instead of two augmentations improve the performance on all metrics apart from the SMA on SpuCo Birds where the student performance goes down but remains within the standard deviation. The experimental setting is consistent with Tab.~\ref{tab:group_shift}.}
    \label{fig:augmentations_scaling}
    %\vspace{-0.05cm}
\end{figure}

In Sec.~\ref{sec:experiments} we used one synthetic augmentation per image using ConfiG and the baselines matched by generation FLOPs according to the multipliers listed in App.~\ref{app:cost}.
In Tab.~\ref{tab:more_aug} we report the results with two ConfiG augmentations and twice as many baseline augmentations as in Sec.~\ref{sec:group_shift_datasets}.
On BAR and SpuCo Birds, using two augmentations per image with ConfiG improves the performance in comparison to Tab.~\ref{tab:group_shift}. On CelebA-HQ the results remain within the standard deviation.
In addition, ConfiG is at least on par with the best baseline with respect to the SMA and outperforms all other methods in GMA and WGA.
In Fig.~\ref{fig:augmentations_scaling}, we compare knowledge distillation with one, two and four ConfiG augmentations per image.
On BAR and CelebA-HQ we observe improvements with four instead of two augmentations per image. On SpuCo Birds, the SMA with four augmentations per image decreases in comparison to two augmentations per image results but remains within the standard deviation.
\begin{table*}[tb]
\centering
\begin{threeparttable}
\caption{\textbf{Full Results with More Augmentations per Image.} Distilling a student with two augmentations per image for ConfiG and all baselines with doubled multipliers from App.~\ref{app:cost} improves the performance in comparison to Tab.~\ref{tab:group_shift} on SpuCo Birds and BAR. All hyperparameters are kept the same, including the replacement probability $\alpha=0.5$. }
\label{tab:more_aug}
\begin{tabular}{@{}lcccccc@{}}
\toprule
& \multicolumn{3}{c}{\textbf{CelebA-HQ}} & \multicolumn{2}{c}{\textbf{SpuCo Birds}} & \textbf{BAR} \\
\cmidrule(lr){2-4} \cmidrule(lr){5-6} \cmidrule(lr){7-7}
\textbf{Method} & \textbf{SMA} & \textbf{GMA} & \textbf{WGA} & \textbf{SMA} & \textbf{WGA} & \textbf{SMA} \\
\midrule

GIF & 97.02\,\stdv{0.45} & 90.79\,\stdv{1.10} & 70.63\,\stdv{4.76} & 63.19\,\stdv{1.63} & 26.17\,\stdv{2.30} & 50.05\,\stdv{3.59} \\

DistDiff & 96.30\,\stdv{0.98} & 89.09\,\stdv{1.10} & 62.83\,\stdv{5.86} & 69.82\,\stdv{2.25} & 24.97\,\stdv{3.33} & 48.09\,\stdv{3.11} \\

DA-Fusion & 97.34\,\stdv{0.49} & 92.14\,\stdv{0.80} & 73.81\,\stdv{4.87} & \underline{71.60}\,\stdv{0.79} & 27.73\,\stdv{4.37} & 56.40\,\stdv{3.99} \\

Diff-Mix & 97.61\,\stdv{0.19} & 93.62\,\stdv{0.44} & 72.62\,\stdv{2.83} & 69.85\,\stdv{2.01} & \underline{34.53}\,\stdv{5.30} & \underline{61.44}\,\stdv{2.11} \\

Diff-II & 94.92\,\stdv{2.67} & 89.41\,\stdv{1.27} & 62.58\,\stdv{3.09} & 60.78\,\stdv{3.33} & 22.67\,\stdv{6.09} & 58.31\,\stdv{2.45} \\

ActGen & \textbf{97.88}\,\stdv{0.24} & \underline{94.72}\,\stdv{0.53} & \underline{84.12}\,\stdv{3.11} & 65.03\,\stdv{1.40} & 25.07\,\stdv{3.13} & 57.31\,\stdv{4.39} \\

ConfiG & \underline{97.75}\,\stdv{0.18} & \textbf{95.67}\,\stdv{0.34} & \textbf{87.83}\,\stdv{1.64} & \textbf{74.64}\,\stdv{1.72} & \textbf{46.90}\,\stdv{6.36} & \textbf{63.12}\,\stdv{1.71} \\

\bottomrule
\end{tabular}
\end{threeparttable}
%\vspace{-.3cm}
\end{table*}

\begin{table*}[tbp]
\centering
\begin{threeparttable}
\caption{\textbf{Different Teacher and Student Models.} Results for a ConvNext V2 Atto distilled from a CLIP ConvNext-B W teacher in the setting of Tab.~\ref{tab:group_shift}. The results of ConfiG are within the standard deviation of the best method or better in SMA and in GMA and WGA it outperforms
all baselines.}
\label{tab:convnext}
\begin{tabular}{@{}lcccccc@{}}
\toprule
& \multicolumn{3}{c}{\textbf{CelebA-HQ}} & \multicolumn{2}{c}{\textbf{SpuCo Birds}} & \textbf{BAR} \\
\cmidrule(lr){2-4} \cmidrule(lr){5-6} \cmidrule(lr){7-7}
\textbf{Method} & \textbf{SMA} & \textbf{GMA} & \textbf{WGA} & \textbf{SMA} & \textbf{WGA} & \textbf{SMA} \\
\midrule
Teacher & 99.10 & 97.94 & 93.24 & 94.90 & 90.40 & 83.79 \\
\midrule
Real Data & 91.02\,\stdv{1.91} & 82.36\,\stdv{2.16} & 42.06\,\stdv{6.11} & 57.98\,\stdv{2.19} & 13.40\,\stdv{5.48} & 39.89\,\stdv{3.88} \\
\midrule
GIF & 93.92\,\stdv{0.88} & 87.44\,\stdv{0.39} & \underline{60.58}\,\stdv{5.30} & 67.82\,\stdv{1.62} & 32.23\,\stdv{3.79} & 51.66\,\stdv{2.48} \\

DistDiff & 92.25\,\stdv{1.07} & 85.59\,\stdv{1.77} & 54.76\,\stdv{5.77} & 67.64\,\stdv{1.08} & 31.57\,\stdv{2.87} & 50.05\,\stdv{2.66} \\

DA-Fusion & 93.96\,\stdv{0.78} & 86.54\,\stdv{0.59} & 56.35\,\stdv{2.13} & \underline{71.35}\,\stdv{1.61} & 32.63\,\stdv{6.07} & 56.42\,\stdv{1.52} \\

Diff-Mix & \underline{96.10}\,\stdv{0.27} & \underline{90.36}\,\stdv{0.64} & 58.60\,\stdv{4.67} & 69.23\,\stdv{2.81} & \underline{33.43}\,\stdv{5.70} & 57.26\,\stdv{0.68} \\

Diff-II & 91.22\,\stdv{1.08} & 84.45\,\stdv{1.06} & 51.32\,\stdv{4.13} & 65.13\,\stdv{1.61} & 27.80\,\stdv{1.91} & \underline{58.82}\,\stdv{1.18} \\

ActGen & 95.86\,\stdv{0.88} & 89.68\,\stdv{0.50} & 57.26\,\stdv{8.22} & 65.60\,\stdv{1.32} & 30.83\,\stdv{2.08} & 56.09\,\stdv{1.51} \\

ConfiG & \textbf{96.35}\,\stdv{0.31} & \textbf{94.19}\,\stdv{0.76} & \textbf{85.20}\,\stdv{3.10} & \textbf{74.97}\,\stdv{2.77} & \textbf{45.60}\,\stdv{5.26} & \textbf{61.52}\,\stdv{1.99} \\

\bottomrule
\end{tabular}
\end{threeparttable}
%\vskip-0.1cm
\end{table*}

\section{Different Student and Teacher Models}\label{app:different_student_models}

For our main experiments in Sec.~\ref{sec:experiments}, we used a ViT-T student~\cite{vit}\footnote{\url{https://huggingface.co/timm/vit_tiny_patch16_224.augreg_in21k}} with 5.8 million parameters and a CLIP ViT-L teacher pre-trained on DataComp XL\footnote{\url{https://huggingface.co/laion/CLIP-ViT-L-14-DataComp.XL-s13B-b90K}}. In Tab.~\ref{tab:convnext}, we report the results for distilling a ConvNext V2 Atto~\cite{convnextv2} student with 3.7M parameters\footnote{\url{https://huggingface.co/timm/convnextv2_atto.fcmae_ft_in1k}} from a CLIP ConvNext-B W~\cite{convnet}\footnote{\url{https://huggingface.co/laion/CLIP-convnext_base_w-laion2B-s13B-b82K}} teacher trained on LAION 2B~\cite{laion}. We keep the same training hyperparameters as for the ViT models. The initial model weights for the student are pre-trained on ImageNet-1k and downloaded using the timm library~\cite{rw2019timm}. We find that also for these additional student and teacher architectures, students distilled with ConfiG consistently achieve SMA scores within the standard deviation of the best baseline or better and outperform all baselines in GMA and WGA.

\section{Different Replacement Probabilities}\label{app:replacement_prob}

DA-Fusion and Diff-II set a fixed replacement probability of $\alpha=0.5$. For our main results in Sec.~\ref{sec:experiments} we follow this setup. In Tab.~\ref{tab:25syn} and Tab.~\ref{tab:75syn} we ablate on the choice of $\alpha$ with $\alpha=0.25$ and $\alpha=0.75$. On SpuCo Birds a smaller $\alpha$ generally decreases performance, in particular the WGA. On CelebA-HQ the variations between different values of $\alpha$ remain within the standard deviation for the majority of methods. For BAR we observe that $\alpha=0.75$ yields noticeable improvements for DA-Fusion, Diff-Mix and DistDiff while the performance of ActGen decreases. Irrespective of the value of $\alpha$, ConfiG achieves the best or second-best performance among all methods.

\begin{table*}[tbp]
\centering
\begin{threeparttable}
\caption{Distilling students using synthetic data augmentations with a replacement probability of $\alpha=0.25$ in the setting of Tab.~\ref{tab:group_shift}. The colored numbers state the difference compared to $\alpha=0.5$ in Tab.~\ref{tab:group_shift}. \textcolor{red}{Red} indicates decreased and \textcolor{green}{green} indicates improved performance.}
\label{tab:25syn}
\begin{tabular}{@{}lcccccc@{}}
\toprule
& \multicolumn{3}{c}{\textbf{CelebA-HQ}} & \multicolumn{2}{c}{\textbf{SpuCo Birds}} & \textbf{BAR} \\
\cmidrule(lr){2-4} \cmidrule(lr){5-6} \cmidrule(lr){7-7}
\textbf{Method} & \textbf{SMA} & \textbf{GMA} & \textbf{WGA} & \textbf{SMA} & \textbf{WGA} & \textbf{SMA} \\
\midrule

GIF & 97.11\,\stdv{0.77} & 90.69\,\stdv{1.24} & 67.19\,\stdv{4.99} & 60.92\,\stdv{1.89} & 21.40\,\stdv{2.44} & 45.57\,\stdv{3.82} \\

&{\scriptsize \textcolor{green}{+0.41}}&{\scriptsize \textcolor{red}{-0.09}}&{\scriptsize \textcolor{red}{-1.72}}&{\scriptsize \textcolor{red}{-2.23}}&{\scriptsize \textcolor{red}{-4.50}}&{\scriptsize \textcolor{red}{-2.34}}\\

DistDiff & 95.62\,\stdv{1.28} & 90.16\,\stdv{1.32} & 68.92\,\stdv{5.80} & 63.88\,\stdv{3.41} & 22.50\,\stdv{3.60} & 45.95\,\stdv{3.91}\\

&{\scriptsize \textcolor{red}{-0.71}}&{\scriptsize \textcolor{red}{-0.25}}&{\scriptsize \textcolor{red}{-1.33}}&{\scriptsize \textcolor{red}{-0.06}}&{\scriptsize \textcolor{red}{-1.57}}&{\scriptsize \textcolor{green}{+0.04}}\\

DA-Fusion & 96.19\,\stdv{0.67} & 90.63\,\stdv{0.64} & 69.97\,\stdv{3.49} & 70.85\,\stdv{1.19} & 30.40\,\stdv{5.97} & 52.45\,\stdv{2.55} \\

&{\scriptsize \textcolor{green}{+0.80}}&{\scriptsize \textcolor{green}{+0.54}}&{\scriptsize \textcolor{green}{+2.01}}&{\scriptsize \textcolor{red}{-0.60}}&{\scriptsize \textcolor{green}{+0.20}}&{\scriptsize \textcolor{red}{-0.59}} \\

Diff-Mix & 97.31\,\stdv{0.31} & 93.07\,\stdv{0.89} & 70.24\,\stdv{4.94} & 65.59\,\stdv{3.50} & 24.57\,\stdv{5.72} & 59.76\,\stdv{2.27} \\

&{\scriptsize \textcolor{red}{-0.05}}&{\scriptsize \textcolor{green}{+0.00}}&{\scriptsize \textcolor{red}{-0.40}}&{\scriptsize \textcolor{red}{-3.90}}&{\scriptsize \textcolor{red}{-7.57}}&{\scriptsize \textcolor{green}{+0.48}} \\

Diff-II  & 94.78\,\stdv{1.00} & 89.28\,\stdv{1.55} & 65.61\,\stdv{8.41} & 59.48\,\stdv{3.07} & 16.90\,\stdv{7.12} & 54.26\,\stdv{3.43} \\

&{\scriptsize \textcolor{red}{-0.30}}&{\scriptsize \textcolor{red}{-0.20}}&{\scriptsize \textcolor{green}{+0.65}}&{\scriptsize \textcolor{red}{-2.40}}&{\scriptsize \textcolor{red}{-6.00}}&{\scriptsize \textcolor{red}{-2.14}} \\

ActGen & 96.55\,\stdv{0.79} & 92.19\,\stdv{1.36} & 76.32\,\stdv{6.91} & 64.37\,\stdv{2.60} & 23.87\,\stdv{3.02} & 52.68\,\stdv{4.08}  \\

&{\scriptsize \textcolor{red}{-0.48}}&{\scriptsize \textcolor{red}{-0.42}}&{\scriptsize \textcolor{green}{+0.64}}&{\scriptsize \textcolor{red}{-0.18}}&{\scriptsize \textcolor{red}{-2.10}}&{\scriptsize \textcolor{red}{-3.03}} \\

ConfiG & 97.94\,\stdv{0.11} & 96.02\,\stdv{0.33} & 89.47\,\stdv{1.91} & 70.23\,\stdv{1.20}& 36.07\,\stdv{4.08}& 58.69\,\stdv{3.89} \\

&{\scriptsize \textcolor{green}{+0.19}}&{\scriptsize \textcolor{green}{+0.36}}&{\scriptsize \textcolor{green}{+1.32}}&{\scriptsize \textcolor{red}{-3.16}}&{\scriptsize \textcolor{red}{-3.43}}&{\scriptsize \textcolor{green}{+0.03}} \\

\bottomrule
\end{tabular}
\end{threeparttable}
%\vspace{-.1cm}
\end{table*}
\begin{table*}[tbp]
\centering
\begin{threeparttable}
\caption{Distilling students using synthetic data augmentations with a replacement probability of $\alpha=0.75$ in the setting of Tab.~\ref{tab:group_shift}. As in Tab.~\ref{tab:25syn}, the colored numbers state the difference compared to $\alpha=0.5$ in Tab.~\ref{tab:group_shift}.}
\label{tab:75syn}
\begin{tabular}{@{}lcccccc@{}}
\toprule
& \multicolumn{3}{c}{\textbf{CelebA-HQ}} & \multicolumn{2}{c}{\textbf{SpuCo Birds}} & \textbf{BAR} \\
\cmidrule(lr){2-4} \cmidrule(lr){5-6} \cmidrule(lr){7-7}
\textbf{Method} & \textbf{SMA} & \textbf{GMA} & \textbf{WGA} & \textbf{SMA} & \textbf{WGA} & \textbf{SMA} \\
\midrule

GIF & 96.85\,\stdv{0.86} & 91.34\,\stdv{1.06} & 71.03\,\stdv{3.43} & 64.64\,\stdv{2.05} & 28.00\,\stdv{4.12} & 46.74\,\stdv{2.94} \\

&{\scriptsize \textcolor{green}{+0.15}}&{\scriptsize \textcolor{green}{+0.56}}&{\scriptsize \textcolor{green}{+2.12}}&{\scriptsize \textcolor{green}{+1.49}}&{\scriptsize \textcolor{green}{+2.10}}&{\scriptsize \textcolor{red}{-1.17}}\\

DistDiff & 96.62\,\stdv{0.43} & 90.08\,\stdv{0.82} & 65.74\,\stdv{6.16} & 65.98\,\stdv{2.04} & 22.50\,\stdv{3.59} & 49.29\,\stdv{2.69}\\

&{\scriptsize \textcolor{green}{+0.29}}&{\scriptsize \textcolor{red}{-0.33}}&{\scriptsize \textcolor{red}{-4.51}}&{\scriptsize \textcolor{green}{+2.04}}&{\scriptsize \textcolor{red}{-1.57}}&{\scriptsize \textcolor{green}{+3.38}}\\

DA-Fusion & 97.07\,\stdv{0.48} & 92.14\,\stdv{0.47} & 74.97\,\stdv{5.35} & 71.14\,\stdv{2.22} & 30.46\,\stdv{4.87} & 56.88\,\stdv{4.44} \\

&{\scriptsize \textcolor{green}{+1.68}}&{\scriptsize \textcolor{green}{+2.05}}&{\scriptsize \textcolor{green}{+7.01}}&{\scriptsize \textcolor{red}{-0.31}}&{\scriptsize \textcolor{green}{+0.26}}&{\scriptsize \textcolor{green}{+3.85}} \\

Diff-Mix & 97.59\,\stdv{0.14} & 94.09\,\stdv{0.48} & 75.40\,\stdv{3.33} & 70.41\,\stdv{2.39} & 35.67\,\stdv{7.60} & 62.05\,\stdv{2.40} \\

&{\scriptsize \textcolor{green}{+0.23}}&{\scriptsize \textcolor{green}{+1.02}}&{\scriptsize \textcolor{green}{+4.76}}&{\scriptsize \textcolor{green}{+0.92}}&{\scriptsize \textcolor{green}{+3.53}}&{\scriptsize \textcolor{green}{+2.78}} \\

Diff-II  & 95.27\,\stdv{0.81} & 88.51\,\stdv{1.34} & 57.01\,\stdv{6.49} & 62.02\,\stdv{2.40} & 24.20\,\stdv{5.45} & 55.86\,\stdv{1.32} \\

&{\scriptsize \textcolor{green}{+0.18}}&{\scriptsize \textcolor{red}{-0.96}}&{\scriptsize \textcolor{red}{-7.95}}&{\scriptsize \textcolor{green}{+0.14}}&{\scriptsize \textcolor{green}{+1.30}}&{\scriptsize \textcolor{red}{-0.54}} \\

ActGen & 97.12\,\stdv{0.35} & 92.89\,\stdv{0.92} & 77.65\,\stdv{4.27} & 64.67\,\stdv{1.82} & 26.50\,\stdv{1.65} & 52.19\,\stdv{5.07} \\

&{\scriptsize \textcolor{green}{+0.09}}&{\scriptsize \textcolor{green}{+0.29}}&{\scriptsize \textcolor{green}{+1.96}}&{\scriptsize \textcolor{green}{+0.13}}&{\scriptsize \textcolor{green}{+0.53}}&{\scriptsize \textcolor{red}{-3.52}} \\

ConfiG & 97.76\,\stdv{0.11} & 95.72\,\stdv{0.36} & 88.77\,\stdv{1.47} & 73.50\,\stdv{1.39} & 44.90\,\stdv{3.60} & 58.61\,\stdv{3.07} \\

&{\scriptsize \textcolor{green}{+0.00}}&{\scriptsize \textcolor{green}{+0.06}}&{\scriptsize \textcolor{green}{+0.62}}&{\scriptsize \textcolor{green}{+0.12}}&{\scriptsize \textcolor{green}{+5.40}}&{\scriptsize \textcolor{red}{-0.05}} \\

\bottomrule
\end{tabular}
\end{threeparttable}
\end{table*}

\section{Teacher-Student Agreement and Loss on Synthetic Data Augmentations}\label{app:aug_loss}

Tab.~\ref{tab:aug_loss} reports the teacher-student agreement and EDRM loss on the synthetic data augmentations, using the auxiliary student used for generating ConfiG augmentations. 
We find that on the samples generated by the baseline the teacher and student agree more often and the EDRM loss is lower than for ConfiG.
Samples generated with ConfiG consistently yield the lowest student accuracy and highest EDRM loss.
This indicates that ConfiG successfully alters spurious features which the student relies on (due to missing groups in the training data) more often than the baselines.

\begin{table*}[tbp]
\centering
\begin{threeparttable}
\caption{\textbf{Teacher-Student Agreement and Loss on Synthetic Data Augmentations.} On ConfiG augmentations the teacher-student agreement, denoted by \textbf{Agree.} in $\%$, is lower and the EDRM loss $R^D$ is higher in comparison to the baselines.}
\label{tab:aug_loss}
\begin{tabular}{@{}lcccccc@{}}
\toprule
& \multicolumn{2}{c}{\textbf{CelebA-HQ}} & \multicolumn{2}{c}{\textbf{SpuCo Birds}} & \multicolumn{2}{c}{\textbf{BAR}} \\
\cmidrule(lr){2-3} \cmidrule(lr){4-5} \cmidrule(lr){6-7}
\textbf{Method} & \textbf{Agree.}$\downarrow$ & \textbf{$R^D$}$\uparrow$ & \textbf{Agree.}$\downarrow$ & \textbf{$R^D$}$\uparrow$ & \textbf{Agree.}$\downarrow$ & \textbf{$R^D$}$\uparrow$ \\
\midrule
GIF & \underline{68.35} & \underline{0.90} & \underline{75.65} & 0.66 & \underline{60.72} & \underline{1.37} \\
DistDiff & 96.18 & 0.12 & 93.20 & 0.34 & 97.93 & 0.08 \\
DA-Fusion & 92.37 & 0.20 & 86.05 & 0.71 &81.95 & 0.77 \\
Diff-Mix & 93.40 & 0.24 & 90.71 & 0.44 & 82.36 & 0.89 \\
Diff-II & 99.48 & 0.28 & 97.42 & 0.18 & 81.56 & 0.75 \\
ActGen & 72.48 & 0.66 & 78.55 & \underline{0.86} & 66.38 & 1.16 \\
ConfiG & \textbf{20.00} & \textbf{2.77} & \textbf{37.60} & \textbf{2.56} & \textbf{20.33} & \textbf{3.08} \\

\bottomrule
\end{tabular}
\end{threeparttable}
\end{table*}

\section{Full Results with Weaker Teacher Models}\label{app:weaker_teacher}

To investigate the effect of using weaker teacher models, we use a RN50 trained on YFCC15M\footnote{\url{https://huggingface.co/timm/resnet50_clip.yfcc15m}} as teacher for the CelebA-HQ dataset and a CLIP ViT-B/32 trained on DataComp M\footnote{\url{https://huggingface.co/laion/CLIP-ViT-B-32-DataComp.M-s128M-b4K}} as teacher for the BAR dataset. The student model and training hyperparameters are kept as in Sec.~\ref{sec:group_shift_datasets}. Tab.~\ref{tab:weaker_teacher_tab} demonstrates that a weaker teacher can decrease the student performance, though ConfiG still improves the performance over real data and achieves the best or second-best student performances.

\begin{table*}[tbp]
\centering
\begin{threeparttable}
\caption{\textbf{Results Using Weaker Teacher Models.} We replace the CLIP ViT-L/14 used in Sec.~\ref{sec:experiments} by weaker teachers (RN50 for CelebA-HQ and CLIP ViT-B/32 for BAR). This decreases the student performance but ConfiG still improves the performance over only real data and reaches the best or second-best student performances.}
\label{tab:weaker_teacher_tab}
\begin{tabular}{@{}lcccc@{}}
\toprule
& \multicolumn{3}{c}{\textbf{CelebA-HQ}} & \multicolumn{1}{c}{\textbf{BAR}} \\
\cmidrule(lr){2-4} \cmidrule(lr){5-5}
\textbf{Method} & \textbf{SMA} & \textbf{GMA} & \textbf{WGA} & \textbf{SMA}  \\
\midrule
Teacher & 96.48 & 92.23 & 75.68 & 59.63 \\
\midrule
Real Data &
94.37\stdv{0.47} & 83.37\stdv{1.48} & 38.36\stdv{5.37} &
36.65\stdv{1.68}  \\
\midrule
GIF &
94.94\stdv{0.60} & 85.89\stdv{0.74} & 48.41\stdv{3.17} &
39.27\stdv{1.14}  \\
DistDiff &
95.08\stdv{0.46} & 85.83\stdv{1.00} & 49.47\stdv{3.84} &
38.86\stdv{1.29}  \\
DA-Fusion &
\textbf{95.27}\stdv{0.54} & 86.54\stdv{1.17} & \underline{56.08}\stdv{4.28} &
38.63\stdv{2.03}  \\
Diff-Mix &
94.48\stdv{0.73} & \underline{86.62}\stdv{1.33} & 50.79\stdv{4.99} &
40.67\stdv{2.82}  \\
Diff-II &
94.61\stdv{0.58} & 85.36\stdv{1.32} & 45.63\stdv{4.13} &
\underline{43.81}\stdv{1.82}  \\
ActGen &
94.84\stdv{0.48} & 85.43\stdv{0.85} & 47.35\stdv{3.75} &
40.01\stdv{1.69}  \\
ConfiG &
\underline{95.25}\stdv{0.52} & \textbf{90.16}\stdv{0.63} & \textbf{67.72}\stdv{2.91} &
\textbf{44.55}\stdv{1.76}  \\
\bottomrule
\end{tabular}
\end{threeparttable}
\end{table*}

\section{Larger Number of Images per Class for SpuCo Birds}\label{app:larger_trainingset}

For Tab.~\ref{tab:results_spuco_larger} we repeat the results for SpuCo Birds from Sec.~\ref{sec:experiments} with 750 instead of 250 images per class. Per class we sample 90$\%$ of the images such that the landbirds are shown in front of land backgrounds and waterbirds are shown in front of water backgrounds as in Sec.~\ref{sec:group_shift_datasets}. The remaining 10$\%$ of the real training samples conflict with the bias, i.e. landbirds with water background or waterbirds with land background. The training setup is otherwise identical to that used for Tab.~\ref{tab:group_shift} and we compare ConfiG against Diff-Mix and ActGen. With the larger training set and bias-conflicting samples, the performance gap between ConfiG and the baselines narrows. Diff-Mix and ConfiG achieve comparable SMA and WGA while ActGen performs approximately 1$\%$ worse than both methods. ConfiG achieves the highest WGA and slightly outperforms the second-best method Diff-Mix by approximately 2$\%$. However, this difference remains within the standard deviation.

\begin{table*}[bp]
\centering
\begin{threeparttable}
\caption{\textbf{Larger Number of Images per Class.} We ablate the results on SpuCo Birds from Tab.~\ref{tab:group_shift} with a larger training dataset that features 750 instead of 250 images per class. Per class, 10$\%$ of the real training samples conflict with the background bias. The remaining experimental setup including class-level spurious features and the student and teacher model architecture are consistent with Tab.~\ref{tab:group_shift} as described in Sec.~\ref{sec:experiments}. We observe that also for the larger training dataset ConfiG outperforms the baselines.}
\label{tab:results_spuco_larger}
\begin{tabular}{@{}lcc@{}}
\toprule
& \multicolumn{2}{c}{\textbf{SpuCo Birds Larger Trainset}} \\
\cmidrule(lr){2-3}
\textbf{Method \hspace{.9cm}} & \hspace{.9cm} \textbf{SMA}\hspace{.9cm}  & \hspace{.9cm} \textbf{WGA} \hspace{.9cm} \\
\midrule
Real Data & 74.10\,\stdv{1.61} & 40.00\,\stdv{4.69} \\
\midrule
Diff-Mix & \textbf{79.88}\,\stdv{1.24} & \underline{57.63}\,\stdv{4.26} \\
ActGen & 78.43\,\stdv{1.14} & 55.47\,\stdv{2.58} \\
ConfiG & \underline{79.67}\,\stdv{1.22} & \textbf{59.53}\,\stdv{3.83} \\
\bottomrule
\end{tabular}
\end{threeparttable}
\end{table*}

\section{Training Data Without Covariate Shift}\label{app:pets}

While the main focus of ConfiG is to improve knowledge distillation under covariate shift, its applicability is not limited to such settings. We demonstrate this in a 5-shot setting on the Oxford Pets dataset~\cite{oxfordpets}, without dedicated covariate shift between the train and test data. 
The results are shown in Tab.~\ref{tab:pets}.
We keep the same teacher and student models as in Sec.~\ref{sec:group_shift_datasets} and train for 300 epochs with batchsize 64 and learning rate 1e-4. The replacement probability is set to $\alpha=0.5$.
As in all our experiments, we use multipliers for the baselines to approximately match the generation cost. For Diff-Mix we generate 20 times as many augmentations as with ConfiG and for ActGen we generate 6 times as many augmentations as ConfiG. Due to the specialized names of the pet classes we add the suffix ``which is a type of pet'' to the prompt for ConfiG. We observe that with one ConfiG augmentation per image, Diff-Mix and ActGen substantially outperform ConfiG. When scaling ConfiG to 8 augmentations per image and multiplying the number of augmentation from the baselines accordingly, we observe that the student performance with ConfiG is comparable to Diff-Mix.

\begin{table*}[tbp]
\centering
\begin{threeparttable}
\caption{\textbf{Distillation Without Covariate Shift.} We keep the same student and teacher model for distillation on a 5-shot subset of the Oxford Pets dataset~\cite{oxfordpets}. When using a single ConfiG augmentation per image, ActGen and Diff-Mix substantially outperform ConfiG. When scaling ConfiG to 8 augmentations per image, ConfiG and Diff-Mix achieve comparable performance.}

\begin{tabular}{@{}lc@{}}
\toprule
Method & \textbf{SMA} \\
\midrule

Real Data \hspace{1.4cm} & \hspace{1.4cm}  34.82\stdv{4.33} \hspace{1.4cm}  \\

\midrule
\multicolumn{2}{c}{\textit{1$\times$ ConfiG Augmentations}} \\
\midrule
Diff-Mix \hspace{1.4cm} & \hspace{1.4cm}  \textbf{71.04}\stdv{2.96} \hspace{1.4cm}  \\
ActGen \hspace{1.4cm}  & \hspace{1.4cm}  60.18\stdv{4.76} \hspace{1.4cm}  \\
ConfiG \hspace{1.4cm}  & \hspace{1.4cm}  50.88\stdv{4.72} \hspace{1.4cm} \\

\midrule
\multicolumn{2}{c}{\textit{8$\times$ ConfiG Augmentations}} \\
\midrule

Diff-Mix \hspace{1.4cm} & \hspace{1.4cm}  74.55\stdv{3.81} \hspace{1.4cm} \\
ActGen \hspace{1.4cm}  & \hspace{1.4cm}  69.76\stdv{2.36} \hspace{1.4cm}  \\
ConfiG \hspace{1.4cm}  & \hspace{1.4cm}  \textbf{74.62}\stdv{4.16}\hspace{1.4cm}  \\

\bottomrule
\end{tabular}
\label{tab:pets}
\end{threeparttable}
\end{table*}

\section{Choosing the Target Class}\label{app:target_class}

\begin{figure}[bt]
    \centering
    \hspace{-.5cm}
    \includegraphics[width=.7\columnwidth]{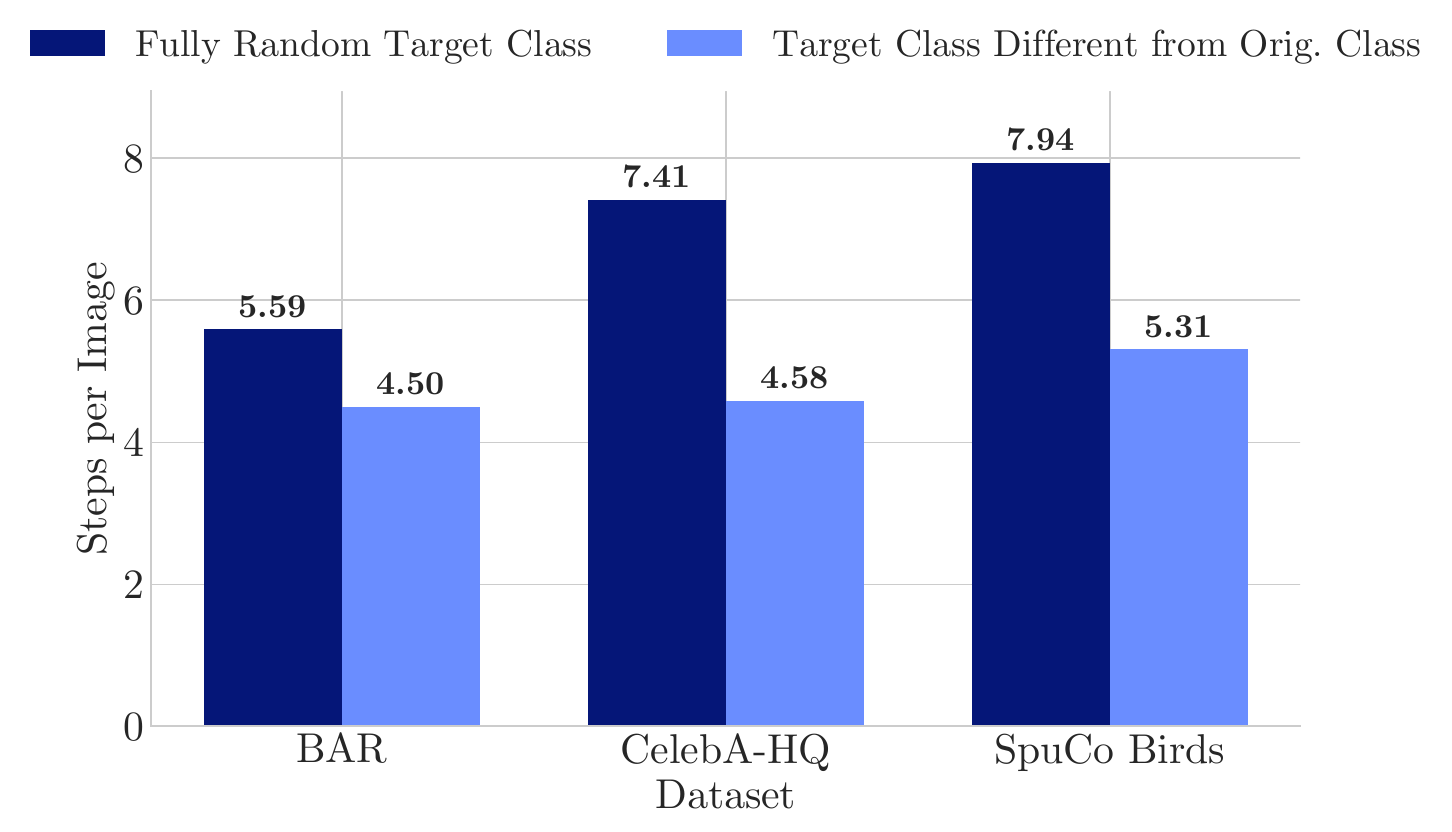}
    %\vskip-0.2cm
    \caption{\textbf{Choosing the Target Class.} Constraining the target class to be different from the original class reduces the average number of steps for ConfiG augmentations per dataset.
    }
    \label{fig:steps_per_im_comparison}
    %\vskip-0.2cm
\end{figure}

As introduced in Sec.~\ref{sec:config}, ConfiG uses a randomly sampled target class which is constrained to be \textit{different} from the original class.
Fig.~\ref{fig:steps_per_im_comparison} shows that using a target class which is different from the original class reduces the number of steps per augmentation on the group shift datasets, in particular for CelebA-HQ and SpuCo Birds where the target class in fact becomes deterministic.
This indicates that the latents and conditioning after inter-class mixing provide a better initialization for the optimization problem defined by Equation~\eqref{eq:loss} if the target class is constrained to be different from the original class.
Tab.~\ref{tab:target_class} shows that choosing a target class which is different from the original class improves the student performance on CelebA-HQ and SpuCo Birds in comparison to a fully random target class that may coincide with the original class. On BAR the performance remains comparable, yet the cost for generating an augmentation decreases due to the lower number of steps required.

\begin{table*}[tbp]
\centering
\begin{threeparttable}
\caption{\textbf{Choosing the Target Class.} Constraining the target class of ConfiG to be different from the original class improves student performance on CelebA-HQ and SpuCo Birds. On BAR the performance is comparable to using a fully random target class but the cost per augmentation is reduced due to the lower number of steps shown in Fig.~\ref{fig:steps_per_im_comparison}.}
\label{tab:target_class}
\begin{tabular}{@{}lcccccc@{}}
\toprule
\textbf{Target} & \multicolumn{3}{c}{\textbf{CelebA-HQ}} & \multicolumn{2}{c}{\textbf{SpuCo Birds}} & \textbf{BAR} \\
\cmidrule(lr){2-4} \cmidrule(lr){5-6} \cmidrule(lr){7-7}
\textbf{Class} & \textbf{SMA} & \textbf{GMA} & \textbf{WGA} & \textbf{SMA} & \textbf{WGA} & \textbf{SMA} \\
\midrule
\footnotesize{Fully}  & & & & & & \\
Random & 97.31\,\stdv{0.27} & 94.11\,\stdv{0.81} & 83.86\,\stdv{2.44} & 68.22\,\stdv{0.86} & 32.00\,\stdv{3.33} & \textbf{58.69}\,\stdv{1.34} \\
\midrule
\footnotesize{Target$\neq$}  & & & & & & \\
Orig. class & \textbf{97.76}\,\stdv{0.25} & \textbf{95.67}\,\stdv{0.55} & \textbf{88.15}\,\stdv{2.10} & \textbf{73.38}\,\stdv{1.28} & \textbf{39.50}\,\stdv{4.53} & 58.66\,\stdv{3.76} \\
\bottomrule
\end{tabular}
\end{threeparttable}
%\vspace{-.3cm}
\end{table*}

\section{Summary of the baselines}\label{app:baseline summary}

In the following, we provide brief descriptions of the baseline methods. For hyperparameters such as guidance scale or the number of denoising steps as well as method-specific parameters, we use the default values in the codebase of every baseline unless stated otherwise.

\paragraph{GIF}~\cite{gid} leverages random perturbations of the latent variables together with a class-based informative score during generation. In addition, the framework leverages custom prompts to enhance diversity. 
We use GIF-SD with the hyperparameters scale=50, strength=0.9 and constraint=0.8 on ImageNet-100. For the domain-specific group shift datasets, we use a lower strength of 0.5.

\paragraph{DistDiff}~\cite{distdiff} constructs hierarchical prototypes to approximate the real data distribution. The latent data points within the diffusion model are optimized to stay close to these prototypes with hierarchical energy guidance.

\paragraph{DA-Fusion}~\cite{dafusion} performs data augmentation by adapting the diffusion model using textual inversion and introducing noise at a diffusion timestep that is randomly sampled from a pre-defined distribution.

\paragraph{Diff-Mix}~\cite{Diff-Mix} uses a fine-tuned diffusion model to perform inter-class augmentation where the conditioning prompts are constructed by randomly sampling one of the classes. For the translation strength we set the sampling strategy in the codebase to ``uniform'' which randomly draws $s \in \{0.3,0.5,0.7,0.9\}$.

\paragraph{Diff-II}~\cite{diff-ii} uses inversion interpolation together with a two-stage denoising that leverages suffix prompts generated by a captioning and a language model for additional context in the prompts. For the SpuCo Birds dataset we re-use suffixes that Diff-II generated for a fine-grained bird classification problem (CUB~\cite{cub} 10-shot). For CelebA-HQ, BAR and ImageNet-100, we generate 10 suffixes per dataset using the Diff-II codebase with a Qwen3 32B model~\cite{qwen3} for summarization. We set the generation hyperparameters to the values used in the 5-shot setting of Diff-II.

\paragraph{ActGen}~\cite{actgen} is the only baseline that can consider the outputs of a downstream model to perform data augmentation. The generation process optimizes a loss to achieve difficulty and diversity together with guidance provided by the attention maps of the downstream model. For our experiments, adversarial samples where the cross-entropy loss is part of the overall loss are generated with probability 0.5, while the remaining samples omit the adversarial loss. To ensure consistency with the ActGen setup, the model used to generate ActGen augmentations is trained with ERM.

\section{Example Augmentations}\label{app:further_examples}

\subsubsection{Failure Cases} 
\begin{figure}[tb]
    \centering
    \hspace{-.5cm}
    \includegraphics[width=.95\columnwidth]{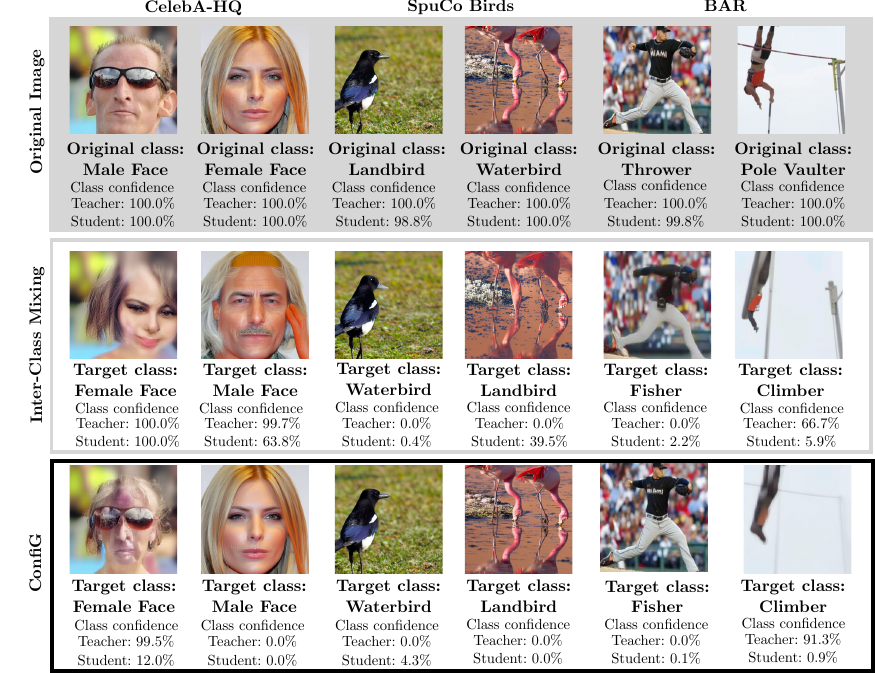}
    \vskip-0.2cm
    \caption{Examples for failure cases of ConfiG data augmentations.}
    \label{fig:failures}
    \vskip-0.2cm
\end{figure}

We categorize the failure cases for ConfiG augmentations into two classes: (1) the augmentations are out-of-distribution images (2) the optimization did not achieve to minimize the loss function. In Fig.~\ref{fig:failures}, we show examples for failures of both types. 
We observe that failure cases (1) occur often for classes which are challenging for the diffusion model to generate, such that the image after inter-class mixing is already non-natural and outside of the domain of the starting image. 
Failure case (1) may contribute to the observation that ConfiG does not consistently outperform the baselines on BAR, unlike on the other group shift datasets where it achieves more consistent gains.
The second failure case can happen for images where the teacher and student both have very high or low confidences on the initial image and the gradients of the loss function from Equation~\eqref{eq:loss} are small.
To mitigate the effect of the second failure case, we employ a rejection sampling strategy with the Hyper-SD single-step model and repeat the augmentation process up to three times if the loss does not reach the stopping criterion.
For Stable Diffusion 1.4 we omit this rejection sampling due to the higher cost of generating augmentations (see App.~\ref{app:cost}).

\subsubsection{Further Examples} 
\begin{figure}[tb]
    \centering
    \hspace{-.5cm}
    \includegraphics[width=.95\columnwidth]{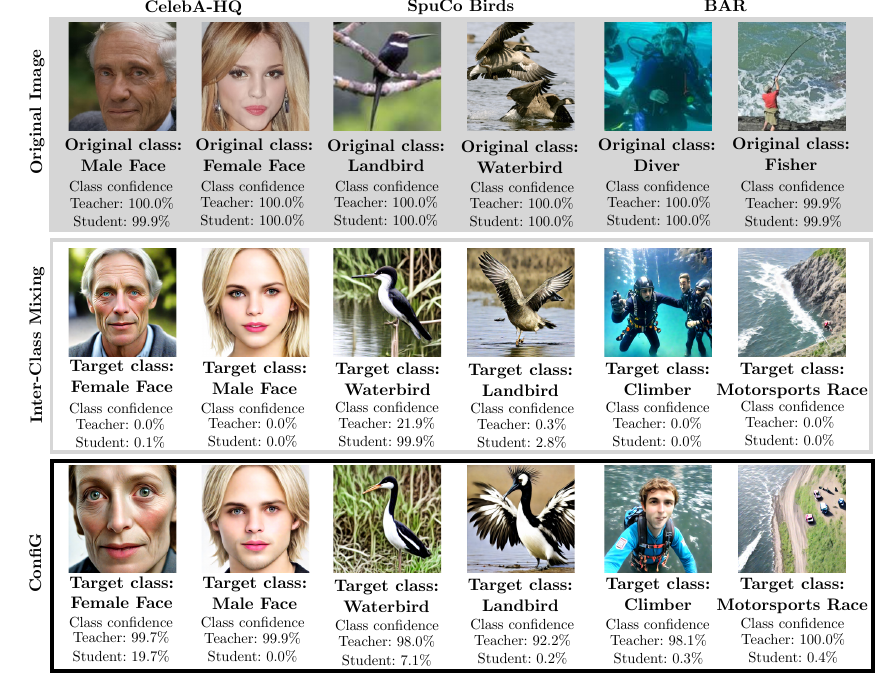}
    \vskip-0.2cm
    \caption{Examples for ConfiG data augmentations with a SDXL Hyper-SD single-step model in the setting described in App.~\ref{sec:hyper_sd}.}
    \label{fig:hyper_sd_examples}
    \vskip-0.2cm
\end{figure}

\begin{figure}[tb]
    \centering
    \hspace{-.5cm}
    \includegraphics[width=.95\columnwidth]{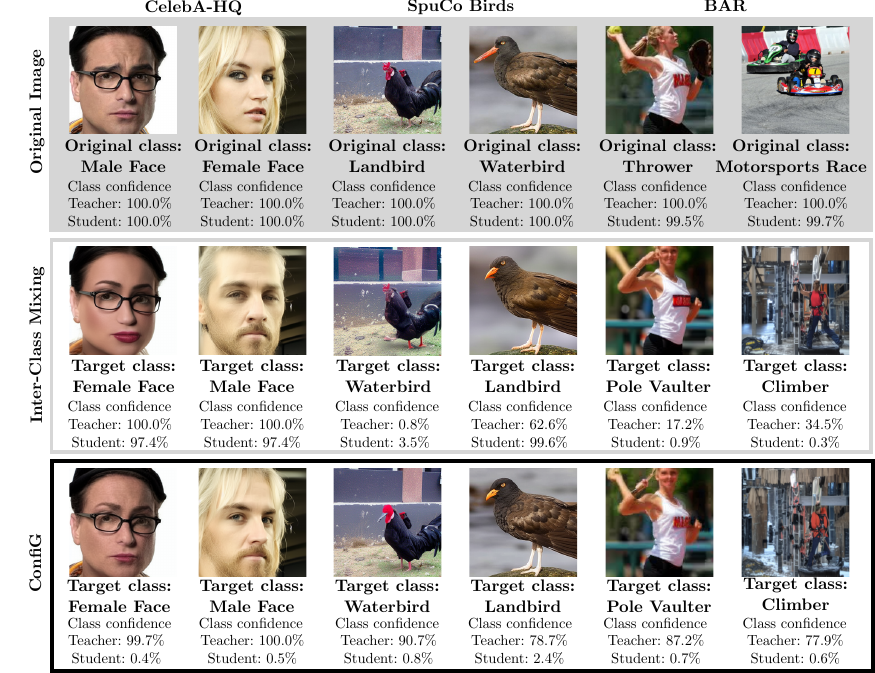}
    \vskip-0.2cm
    \caption{Examples for ConfiG data augmentations for the ConvNext Teacher and student models from App.~\ref{app:different_student_models}.}
    \label{fig:convnext_examples}
    \vskip-0.2cm
\end{figure}

\begin{figure}[tb]
    \centering
    \hspace{-.5cm}
    \includegraphics[width=.95\columnwidth]{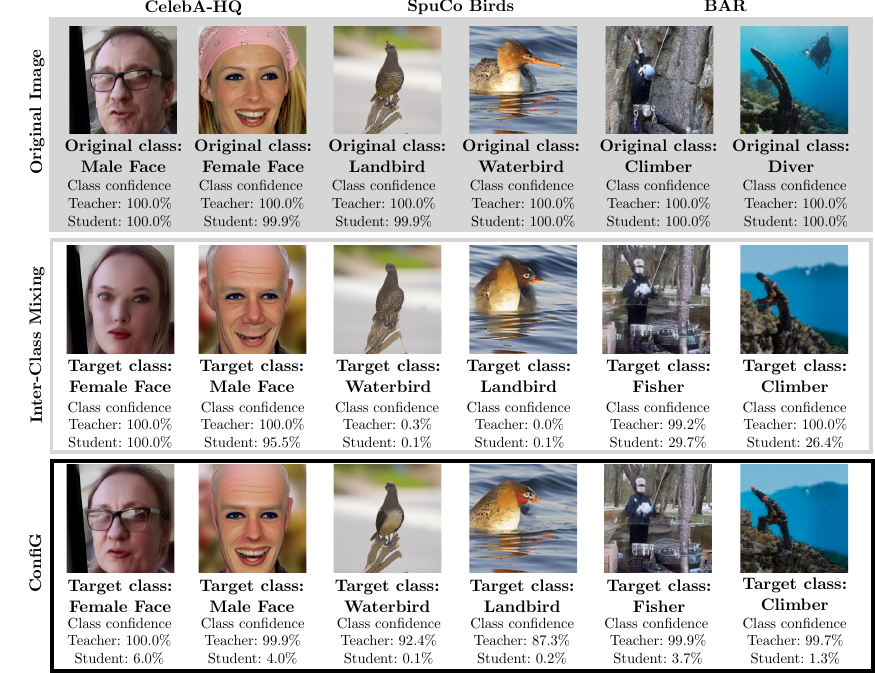}
    \vskip-0.2cm
    \caption{Additional Examples for ConfiG data augmentations used for Tab.~\ref{tab:group_shift}.}
    \label{fig:group_shift_additional}
    \vskip-0.2cm
\end{figure}

\begin{figure}[tb]
    \centering
    \hspace{-.5cm}
    \includegraphics[width=.95\columnwidth]{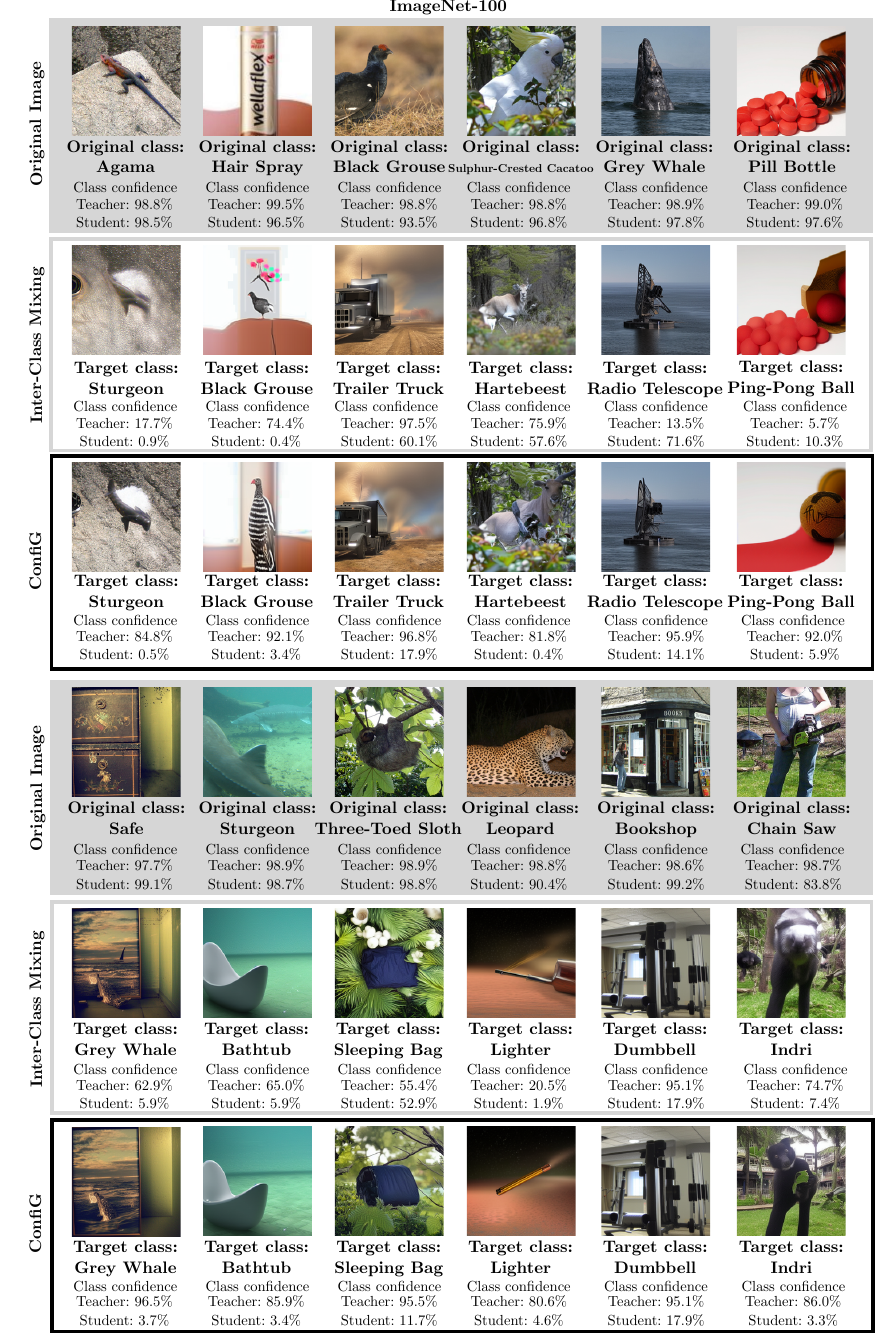}
    \vskip-0.2cm
    \caption{Examples for ConfiG data augmentations from ImageNet-100 with a ViT-T student.}
    \label{fig:imagenet_100_examples}
    \vskip-0.2cm
\end{figure}

Fig.~\ref{fig:hyper_sd_examples} presents images for ConfiG with a single-step diffusion model and IP-Adapter guidance as discussed in App.~\ref{sec:hyper_sd}. In Fig.~\ref{fig:convnext_examples} we show examples for augmentations with the ConvNext teacher and student models from App.~\ref{app:different_student_models}. Additional examples for the group shift datasets from Tab.~\ref{tab:group_shift} in Sec.~\ref{sec:group_shift_datasets} can be found in Fig.~\ref{fig:group_shift_additional}. Fig.~\ref{fig:imagenet_100_examples} shows example augmentations for the ImageNet-100 dataset with a ViT-T student and a BeiTv2-B teacher model. 
We highlight that some of the ConfiG examples find biases of the student which are not actually part of the dataset attributes and were not manually introduced as spurious features in the dataset. In the second example of Fig.~\ref{fig:group_shift_additional}, the ConfiG augmentation shows a male face with eye makeup and a wide smile which likely are correlations we \textit{unknowingly} introduced to the training data.
\newpage

\end{document}